\documentclass{article}

\usepackage{microtype}
\usepackage{graphicx}
\usepackage{booktabs} 
\usepackage{hyperref}

\usepackage{arxiv}

\usepackage{amsmath}
\usepackage{amssymb}
\usepackage{subcaption}

\newcommand{\R}{\mathbb{R}}
\DeclareMathOperator*{\Exp}{\mathbb{E}}
\DeclareMathOperator*{\Var}{\mathbb{V}}
\newcommand{\E}[2][]{\Exp_{#1}\!\left[ #2 \right]}
\newcommand{\V}[2][]{\Var_{#1}\!\left[ #2 \right]}
\newcommand{\calL}{\mathcal{L}}
\newcommand{\diff}[2]{\frac{\partial #1}{\partial #2}}
\newcommand{\norm}[1]{\left\|#1\right\|}
\DeclareMathOperator*{\argmin}{argmin}
\newcommand{\mil}{\emph{Deep MIL}}
\newcommand{\ats}{\emph{attention sampling}}

\icmltitlerunning{Processing Megapixel Images with Deep Attention-Sampling
                  Models}

\begin{document}

\twocolumn[
    \icmltitle{Processing Megapixel Images with Deep Attention-Sampling
               Models}

    \begin{icmlauthorlist}
    \icmlauthor{Angelos Katharopoulos}{idiap,epfl}
    \icmlauthor{Fran\c{c}ois Fleuret}{idiap,epfl}
    \end{icmlauthorlist}

    \icmlaffiliation{idiap}{Idiap Research Institute, Martigny, Switzerland}
    \icmlaffiliation{epfl}{EPFL, Lausanne, Switzerland}

    \icmlcorrespondingauthor{Angelos Katharopoulos}{firstname.lastname@idiap.ch}

    \icmlkeywords{}

    \vskip 0.3in
]

\printAffiliationsAndNotice{}

\begin{abstract}
    Existing deep architectures cannot operate on very large signals such
    as megapixel images due to computational and memory constraints. To
    tackle this limitation, we propose a fully differentiable end-to-end
    trainable model that samples and processes only a fraction of the full
    resolution input image.

    The locations to process are sampled from an attention distribution
    computed from a low resolution view of the input. We refer to our
    method as \emph{attention sampling} and it can process images of
    several megapixels with a standard single GPU setup.

    We show that sampling from the attention distribution results in an
    unbiased estimator of the full model with minimal variance, and we
    derive an unbiased estimator of the gradient that we use to train our
    model end-to-end with a normal SGD procedure.

    This new method is evaluated on three classification tasks, where we
    show that it allows to reduce computation and memory footprint by
    an order of magnitude for the same accuracy as classical
    architectures. We also show the consistency of the sampling that
    indeed focuses on informative parts of the input images.
\end{abstract}

\section{Introduction}

\begin{figure}
    \centering
    \begin{subfigure}[b]{0.68\columnwidth}
      \includegraphics[width=\textwidth]{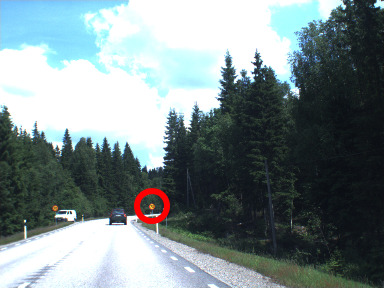}
        \caption{} \label{fig:sign_far:full}
    \end{subfigure}
    \begin{minipage}[b]{0.3\columnwidth}
        \begin{subfigure}[b]{\textwidth}
            \centering
            \includegraphics[width=0.6\textwidth,trim=24 24 30 24]{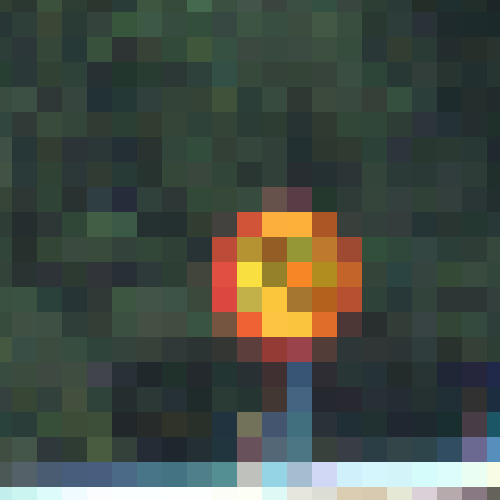}
            \caption{} \label{fig:sign_far:low}
        \end{subfigure}

        \vspace*{1.5ex}

        \begin{subfigure}[b]{\textwidth}
            \centering
            \includegraphics[width=0.6\textwidth,trim=24 24 30 24]{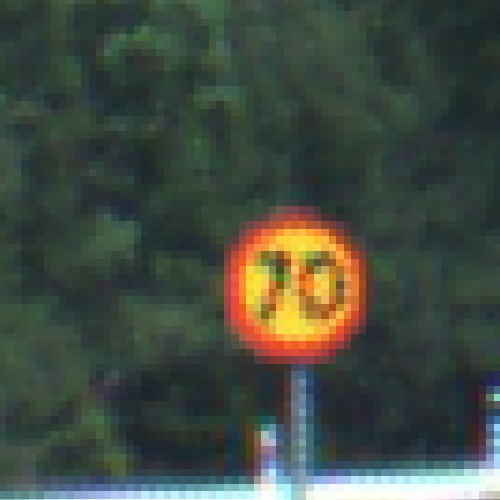}
            \caption{} \label{fig:sign_far:high}
        \end{subfigure}
    \end{minipage}
    \caption{Common practice to process megapixel images with CNNs is to
             downsample them, however this results in significant loss of
             information (b, c).}
    \label{fig:sign_far}
    \vspace{-1.2em}
\end{figure}

For a variety of computer vision tasks, such as cancer detection, self driving
vehicles, and satellite image processing, it is necessary to develop
models that are able to handle high resolution images. The existing CNN
architectures, that provide state-of-the-art performance in various computer vision
fields such as image classification \cite{he2016deep}, object detection
\cite{liu2016ssd}, semantic segmentation \cite{wu2019wider} etc., cannot
operate directly on such images due to computational and memory
requirements. To address this issue, a common practice is to downsample the
original image before passing it to the network. However, this leads to loss of
significant information possibly critical for certain tasks.

Another research direction seeks to mitigate this problem by splitting the
original high resolution image into patches and processing them separately
\cite{hou2016patch, golatkar2018classification, nazeri2018two}.  Naturally,
these methods either waste computational resources on uninformative patches or
require ground truth annotations for each patch. However, per patch labels are
typically expensive to acquire and are not available for the majority of the
available datasets.

The aforementioned limitations are addressed by two disjoint lines of work: the
recurrent visual attention models \cite{mnih2014recurrent, ba2014multiple} and
the attention based multiple instance learning \cite{ilse18a}. The first seeks
to limit the wasteful computations by only processing some parts of the full
image. However, these models result in hard optimization problems that limit
their applicability to high resolution images.  The second line of work shows
that regions of interest can be identified without explicit patch annotations
by aggregating per patch features with an attention mechanism. Nevertheless,
such methods do not address the computational and memory issues inherent in all
patch based models.

This work aims at combining the benefits of both. Towards this goal, we propose
an end-to-end trainable model able to handle multi-megapixel images using a
single GPU or CPU. In particular, we sample locations of ``informative
patches'' from an ``attention distribution'' computed on a lower resolution
version of the original image. This allows us to only process a fraction of the
original image.  Compared to previous works, due to our attention based
weighted average feature aggregation, we are able to derive an unbiased
estimator of the gradient of the virtual and intractable ``full model'' that
would process the full-scale image in a standard feed-forward manner, and do
not need to resort to reinforcement learning or variational methods to train.
%
%
Furthermore, we prove that sampling patches
from the attention distribution results in the minimum variance estimator of
the ``full model''.

We evaluate our model on three classification tasks and we show that our
proposed \ats{} achieves comparable test errors with \citet{ilse18a}, that
considers all patches from the high resolution images, while being up to
\textbf{25$\times$} faster and requiring up to \textbf{30$\times$} less memory.

\section{Related Work}

In this section, we discuss the most relevant body of work on attention-based models
and techniques to process high resolution images using deep neural networks,
which can be trained from a scene-level categorical label.
Region proposal methods that require per-patch annotations, such as
instance-level bounding boxes \cite{girshick2014rich, redmon2016you,
 liu2016ssd}, do not fall in that category.

\subsection{Recurrent visual attention models}

This line of work includes models that learn to extract a sequence of regions
from the original high resolution image and only process these at high
resolution. The regions are processed in a sequential manner, namely the
distribution to sample the $n$-th region depends on the previous $n-1$
regions. \citet{mnih2014recurrent} were the first to employ a recurrent neural
network to predict regions of interest on the high resolution image and process
them sequentially. In order to train their model, which is not differentiable,
they use reinforcement learning. In parallel, \citet{ranzato2014learning,
ba2014multiple} proposed to additionally downsample the input image and use it
to provide spatial context to the recurrent network.
\citet{ramapuram2018variational} improved upon the previous works by using
variational inference and Spatial Transformer Networks
\cite{jaderberg2015spatial} to solve the same optimization problem.

All the aforementioned works seek to solve a complicated optimization problem
that is non differentiable and is approximated with either reinforcement
learning or variational methods. Instead of employing such a complicated model
to aggregate the features of the patches and generate dependent attention
distributions that result in a hard optimization problem, we propose to use an
attention distribution to perform a weighted average of the features, which
allows us to directly train our model with SGD.

\subsection{Patch based models}

Such models \cite{hou2016patch, liu2017detecting, nazeri2018two} divide the
high resolution image into patches and process them separately. Due to the lack
of per patch annotations, the above models need to introduce a separate method
to provide labels for training the patch level network. Instead \ats{} does not
require any patch annotations and through the attention mechanism learns to
identify regions of interest in arbitrarily large images.

\subsection{Attention models}

\citet{xu2015show} were the first to use soft attention methods to generate
image captions.
More related to our work is the model of \citet{ilse18a}, where they use the
attention distribution to aggregate a bag of features. To apply their method to
images, they extract patches, compute features and aggregate them with an
attention distribution that is computed from these features. This allows them
to infer regions of interest without having access to per-patch labels.
However, their model wastes computational resources by handling all patches,
both informative and non-informative. Our method, instead, learns to focus only
on informative regions of the image, thus resulting in orders of magnitude
faster computation while retaining equally good performance.

\subsection{Other methods}

\citet{jaderberg2015spatial} propose Spatial Transformer Networks (STN) that
learn to predict affine transformations of a feature map than includes cropping
and rescaling. STNs employ several localization networks, that operate on the
full image, to generate these transformations. As a result, they do not scale
easily to megapixel images or larger.
\citet{recasens2018learning} use a low resolution view of the image to predict
a saliency map that is used in conjunction with the differentiable STN sampler
to focus on useful regions of the high resolution image by making them larger.
In comparison, \ats{} focuses on regions by weighing the corresponding features
with the attention weights.

\section{Methodology} \label{sec:theory}

In this section, we formalize our proposed \emph{attention-sampling} method.
Initially, we introduce a generic formulation for attention and we show that
sampling from the attention distribution generates an optimal approximation in
terms of variance that significantly reduces the required computation.
In \S~\ref{sec:gradient}, we derive the gradient with respect to
the parameters of the attention and the feature network through the sampling
procedure. In \S~\ref{sec:multires} and \S~\ref{sec:implementation},
we provide the methodology that allows us to speed up the processing of high
resolution images using \ats{} and is used throughout our experiments.

\subsection{Attention in neural networks}

Let $x$, $y$ denote an input-target pair from our dataset. We consider
$\Psi(x; \Theta) = g(f(x; \Theta); \Theta)$ to be a neural network
parameterized by $\Theta$. $f(x; \Theta) \in \R^{K \times D}$ is an
intermediate representation of the neural network that can be thought of as $K$
features of dimension $D$, e.g. the last convolutional layer of a ResNet
architecture or the previous hidden states and outputs of a recurrent neural
network.

Employing an attention mechanism in the neural network $\Psi(\cdot)$ at the
intermediate representation $f(\cdot)$ is equivalent to defining a function $a(x; \Theta)
\in \R_+^K$ s.t. $\sum_{i=1}^K a(x; \Theta)_i = 1$ and changing the definition of the network to
\begin{align} \label{eq:attention}
\Psi(x; \Theta) = g\left(\sum_{i=1}^K a(x; \Theta)_i f(x; \Theta)_i \right),
\end{align}
given that the subscript $i$ extracts the $i$-th row from a matrix or the
$i$-th element from a vector.

\subsection{Attention sampling} \label{sec:att_sampling}

By definition, $a(\cdot)$ is a multinomial distribution over $K$ discrete
elements (e.g.\ locations in the images). Let $I$ be a random variable sampled from $a(x; \Theta)$. We can
rewrite the attention in the neural network $\Psi(\cdot)$ as the expectation of
the intermediate
features over the attention distribution $a(\cdot)$
\begin{align}
\Psi(x; \Theta)
    &= g\left(\sum_{i=1}^K a(x; \Theta)_i f(x; \Theta)_i \right) \\
    &= g\left(\E[I \sim a(x; \Theta)]{f(x; \Theta)_I}\right). \label{eq:attention_expected}
\end{align}

Consequently, we can avoid computing all $K$ features by approximating the
expectation with a Monte Carlo estimate. We sample a set Q of N i.i.d.\ indices
from the attention distribution, $Q = \{q_i \sim a(x; \Theta) \mid i \in \{1, 2,
\dots, N\}\}$ and approximate the neural network with
\begin{align} \label{eq:attention_mc}
\Psi(x; \Theta) \approx
    g\left(
        \frac{1}{N} \sum_{q \in Q} f(x; \Theta)_q
    \right).
\end{align}

\subsubsection{Relation with importance-sampling} \label{sec:is}

We are interested in deriving an approximation with minimum variance so that
the output of the network does not change because of the sampling. In the
following paragraphs, we show that sampling from $a(x; \Theta)$ is optimal in
that respect.

Let $P$ denote a discrete probability distribution on the $K$ features with
probabilities $p_i$. We want to sample from $P$ such that the variance is
minimized. Concretely, we seek $P^*$ such that
\begin{align}
    P^* = \argmin_{P} \V[I \sim P]{\frac{a(x; \Theta)_I f(x; \Theta)_I}{p_I}}.
\end{align}
We divide by $p_I$ to ensure that the expectation remains the same regardless
of $P$. One can easily verify that $\E[I \sim P]{\frac{a(x; \Theta)_I
f(x; \Theta)_I}{p_I}} = \E[I \sim a(x; \Theta)]{f(x; \Theta)_I}$. We continue
our derivation as follows:
\begin{align}
& \argmin_{P} \V[I \sim P]{\frac{a(x; \Theta)_I f(x; \Theta)_I}{p_I}} \\
&\quad = \argmin_{P} \E[I \sim P]{
    \left(\frac{a(x; \Theta)_I}{p_I}\right)^2 \norm{f(x; \Theta)_I}_2^2
    } \\
&\quad = \argmin_{P} \sum_{i=1}^K
    \frac{a(x; \Theta)_i^2}{p_i}\norm{f(x; \Theta)_i}_2^2. \label{eq:is_J}
\end{align}
The minimum of equation \ref{eq:is_J} is
\begin{align}
p_i^* \propto a(x; \Theta)_i \norm{f(x; \Theta)_i}_2,
\end{align}
which means that sampling according to the attention distribution is optimal
when we do not have information about the norm of the features. This can be
easily enforced by constraining the features to have the same $L_2$ norm.

\subsubsection{Gradient derivation} \label{sec:gradient}

In order to use a neural network as our attention distribution we need to
derive the gradient of the loss with respect to the parameters of the attention
function $a(\cdot; \Theta)$ through the sampling of the set of indices $Q$.
Namely, we need to compute
\begin{align} \label{eq:grad_initial}
\diff{\frac{1}{N} \sum_{q \in Q} f(x; \Theta)_q}{\theta}
\end{align}
for all $\theta \in \Theta$ including the ones that affect $a(\cdot)$.

By exploiting the Monte Carlo approximation and the multiply by one trick, we show that
\begin{align}
\diff{}{\theta}\frac{1}{N} \sum_{q \in Q} f(x; \Theta)_q
\approx \E[I \sim a(x; \Theta)]{
    \frac{\diff{}{\theta}\left[a(x; \Theta)_I f(x; \Theta)_I\right]}
        {a(x; \Theta)_I}
    }. \label{eq:grad_final}
\end{align}

In equation \ref{eq:grad_final}, the gradient of each feature is weighed
inversely proportionally to the probability of sampling that feature. This result
is expected, because the ``effect'' of rare samples should be increased to
account for the low observation frequency \cite{kahn1951estimation}.
This allows us to derive the gradients of our \ats{} method as follows:
\begin{align}
\diff{}{\theta}\frac{1}{N} \sum_{q \in Q} f(x; \Theta)_q =
    \frac{1}{N} \sum_{q \in Q}
    \frac{\diff{}{\theta}\left[a(x; \Theta)_q f(x; \Theta)_q\right]}
         {a(x; \Theta)_q},
\end{align}
which requires computing only the rows of $f(\cdot)$ for the sampled indices in
$Q$. Due to lack of space, a detailed derivation of equation
\ref{eq:grad_final}, can be found in our supplementary material.

\subsubsection{Sampling without replacement}

In our initial analysis, we assume that Q is sampled i.i.d.\ from
$a(x; \Theta)$. However, this means that it is probable to sample the same
element multiple times, especially as the entropy of the distribution decreases
during the training of the attention network. To avoid computing a feature
multiple times and to make the best use of the available computational budget
we propose sampling without replacement.

We model sampling without replacement as follows: Initially,
we sample a position $i_1$ with probability $p_1(i) \propto a(x; \Theta)_i \, \forall
i$. Subsequently, we sample the second position $i_2$, given the first, with
probability $p_2(i \mid i_1) \propto a(x; \Theta)_i \, \forall i \neq i_1$. Following
this reasoning, we can define sampling the $n$-th position with probability
\begin{multline}
 \quad \forall i \notin \{i_1, i_2, \dots, i_{n-1}\}, \\
    p_n(i \mid i_1, i_2, \dots, i_{n-1}) \propto a(x; \Theta)_i
\end{multline}
Simply averaging the features, as in equation \ref{eq:attention_mc}, would
result in a biased estimator. Instead, we use
\begin{align}
\Exp_{I_1, I_2, \dots, I_n}\Bigg[ \sum_{k=1}^{n-1} a(x; \Theta)_{I_k} f(x; \Theta)_{I_k} + & \\
    f(x; \Theta)_{I_n}\sum_{t \notin \{I_1, I_2, \dots, I_{n-1}\}} a(x; \Theta)_t & \Bigg] = \\
    \Exp_{I_1, I_2, \dots, I_n}\Bigg[\sum_{i=1}^K a(x; \Theta)_i f(x; \Theta)_i &\Bigg] = \\
    \Exp_{I \sim a(x; \Theta)}\big[ f(x; \Theta)_I &\big].
\end{align}
We assume that $I_1$ to $I_n$ are sampled from $p_1(i)$ to $p_n(i)$
accordingly.
Following the reasoning of \S~\ref{sec:gradient}, we compute the gradient
through the sampling in an efficient and numerically stable way. The complete
analysis is given in the supplementary material.

\subsection{Multi-resolution data} \label{sec:multires}

For most implementations of attention in neural networks, $a(\cdot)$ is a
function of the features $f(\cdot)$ \cite{ilse18a}. This means that in order to
compute the attention distribution we need to compute all the features.
However, in order to take advantage of our Monte Carlo Estimation of equation
\ref{eq:attention_mc} and avoid computing all $K$ features, we use a lower
resolution view of the data. This allows us to gain significant speedup from
\ats{}.

Given an image $x \in \R^{H \times W \times C}$ where $H$, $W$, $C$ denote the
height, width and channels respectively, and its corresponding view $V(x, s)
\in \R^{h \times w \times C}$ at scale $s$ we compute the attention as 
\begin{align}
    a(V(x, s); \Theta) : \R^{h \times w \times C} \to \R^{hw},
\end{align}
where $h < H$ and $w < W$. We also define a function $P(x, i)$ that extracts a
patch from the full resolution image $x$ centered around the corresponding
$i$-th pixel in $V(x, s)$.

Based on the above, we derive a model capable of only
considering few patches from the full size image $x$, as follows:
\begin{align}
\Psi(x) 
    & = g\left(\sum_{i=1}^{hw} a(V(x, s))_i f(P(x, i))\right) \\
    & \approx g\left(\frac{1}{N} \sum_{q \in Q} f(P(x, q))\right).
    \label{eq:mil_pooling}
\end{align}
Note that both the attention and feature functions have trainable
parameters $\Theta$ which we omit for clarity.
In the formulation of equation \ref{eq:mil_pooling}, we do not consider the
location of the sampled patches $P(x, q)$. This is not an inherent limitation
of the model since we can simply pass the location as a parameter in our
feature function $f(\cdot)$.

\subsection{Implementation details} \label{sec:implementation}

In this section, we discuss the specifics of our
proposed \ats{}. Equation $f(\cdot)$ is implemented
by a neural network which we refer to as \emph{feature network}. Similarly
$a(\cdot)$ is another neural network, typically, significantly smaller, referred
to as \emph{attention network}. Finally, function $g(\cdot)$ is a linear
classification layer.

In order to control the exploration-exploitation dilemma we introduce an
entropy regularizer for the attention distribution. Namely given a loss
function $\calL(x, y; \Theta)$ we use
\begin{align}
    \calL'(x, y; \Theta) = \calL(x, y; \Theta) - \lambda \mathcal{H}(a(x; \Theta)),
\end{align}
where $\mathcal{H}(x)$ denotes the entropy of the distribution $x$. This
regularizer prevents the attention network from quickly deciding which
patches are informative. This results in an exploration of the available patch
space during the initial stage of training. Due to lack of space we evaluate
the impact of this regularizer qualitatively in our supplementary.

As already mentioned in \S~\ref{sec:is}, normalizing the features in
terms of the $L_2$ norm guarantees that the attention distribution produces
the minimum variance estimator of the ``full model''; thus in all the feature
networks we add $L_2$ normalization as the final layer.

\begin{figure}
    \centering
    \begin{subfigure}[t]{\columnwidth}
        \centering
        \includegraphics[width=0.8\columnwidth]{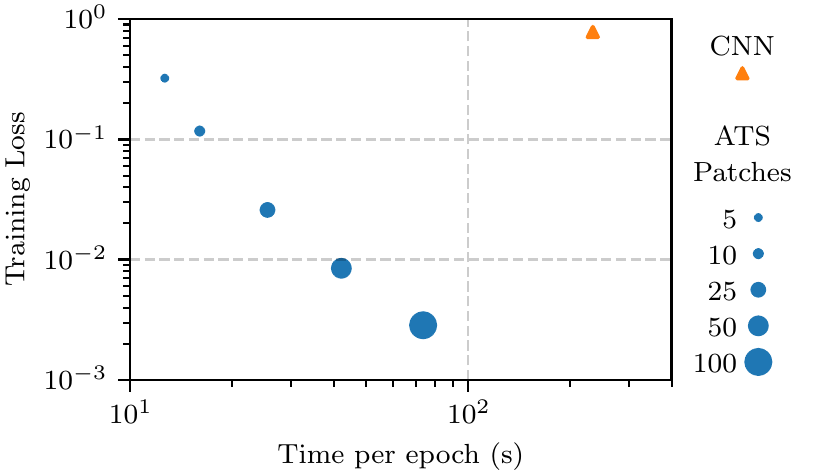}
        \caption{Performance on training set}
        \label{fig:megamnist_performance:a}
    \end{subfigure}
    \begin{subfigure}[t]{\columnwidth}
        \centering
        \includegraphics[width=0.8\columnwidth]{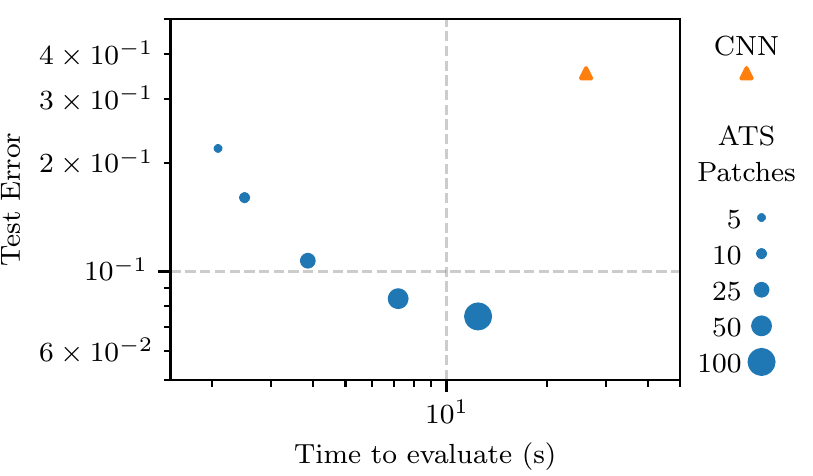}
        \caption{Performance on test set}
        \label{fig:megamnist_performance:b}
    \end{subfigure}
    \caption{Comparison of \ats{} (ATS) with a CNN on Megapixel MNIST. We
             observe that we can trade optimization accuracy for time by
             sampling fewer patches.}
    \label{fig:megamnist_performance}
    \vspace{-1.2em}
\end{figure}

\section{Experimental evaluation}

In this section, we analyse experimentally the performance of our \ats{} approach
on three classification tasks. We showcase the ability of our model to focus
on informative parts of the input image which results in significantly reduced
computational requirements. We refer to our approach as \emph{ATS} or
\emph{ATS-XX} where XX denotes the number of sampled patches. Note that we do
not consider per-patch annotations for any of the used datasets. The code used
for the experiments can be found in
\url{https://github.com/idiap/attention-sampling}.

\subsection{Introduction}

\subsubsection{Baselines}

\begin{figure}
    \centering
    \includegraphics[width=0.5\textwidth]{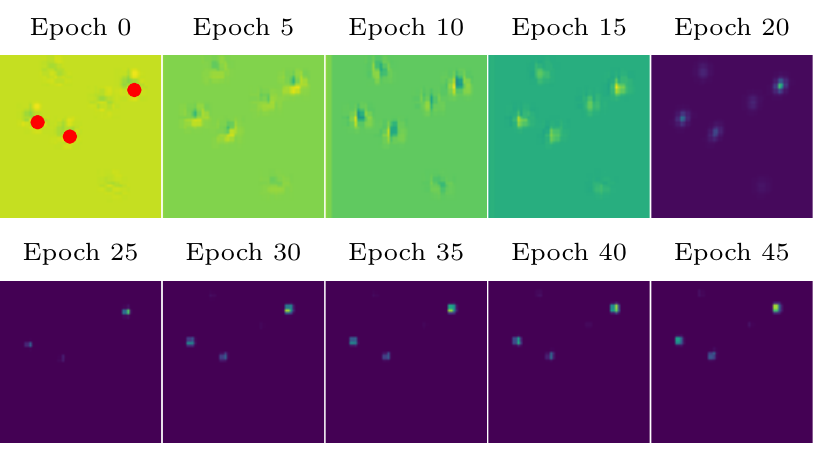}
    \caption{The evolution of the attention distribution on Megapixel MNIST.
    Yellow means higher attention. At the first image (epoch 0) we mark the
    position of the digits with the red dots. The attention finds the three digits and
    focuses on them instead of the noise which can be clearly seen
    in epochs 10 and 15.}
    \label{fig:megamnist_attention}
    \vspace{-1.2em}
\end{figure}

Most related to our method is the patch based approach of \citet{ilse18a} that
implements the attention as a function of the features of each patch. For the
rest of the experiments, we refer to this method as \mil{}. For \mil{}, we specify
the patches to be extracted from each high resolution image by a regular
grid of varying size depending on the dimensions of the input image and the
patch size. Note that the architecture of the feature network and the
patch size used for \mil{} is always the same as the ones used for our \ats{}
method.

To showcase that existing CNN architectures are unable to operate on megapixel
images, we also compare our method to traditional CNN models. Typically, the
approach for handling high resolution images with deep neural networks is to
downsample the input images. Thus; for a fair comparison, we train the CNN
baselines using images at various scales. The specifics of each network
architecture are described in the corresponding experiment. For more details,
we refer the reader to our supplementary material.

Finally, to show that the learned attention distribution is non-trivial, we
replace the attention network of our model with a fixed network that predicts
the uniform distribution and compare the results. We refer to this baseline as
\emph{U-XX} where XX denotes the number of sampled patches.

\subsubsection{Metrics}

Our proposed model allows us to trade off computation with increased
performance. Therefore, besides reporting just the achieved test error, we also
measure the computational and memory requirements. To this end, we report the
per sample wall-clock time for a forward/backward pass and the peak GPU memory
allocated for training with a batch size of 1, as reported by the TensorFlow
\cite{abadi2016tensorflow} profiler. Note that for \ats{}, extracting a patch,
reading it from main memory and moving it to the GPU memory is always included
in the reported time. Regarding the memory requirements of our baselines, it
is important to mention that the maximum used memory depends on the size of the
high resolution image, whereas for \ats{} it only depends on the number sampled
patches and the patch size. For a fair comparison in terms of both memory and
computational requirements, with \mil{}, we make sure that the patches are
extracted from a grid with a stride at least half the size of the patch.
Finally, due to lack of space, we provide extensive qualitative results of the
learned attention distribution in the supplementary material.

\subsection{Megapixel MNIST} \label{sec:megamnist}

We evaluate \ats{} on an artificial dataset based on the MNIST
digit classification task \cite{lecun2010mnist}. We generate $6000$ empty
images of size $1500 \times 1500$ and we place patches of random noise at $50$
random locations. The size of each patch is equal to an MNIST digit. In
addition, we randomly position 5 digits sampled from the MNIST dataset, 3
belonging to the same class and 2 to a random class. The task is to identify
the digit with the most occurrences. We use $5000$ images for training and $1000$
for testing. 


For ATS, the attention network is a three layer convolutional network and the
feature network is inspired from LeNet-1 \cite{lecun1995comparison}. To compute
the attention, we downsample the image to $180 \times 180$ which results in
$32,400$ patches to sample from. The sampled patches from the high resolution
image have size $50\times50$ pixels.  For the CNN baseline, we train it on the
full size images. Regarding uniform sampling, we note that it does not perform better than random
guessing, due to the very large sampling space ($32,400$ possible patches);
thus we omit it from this experiment. Furthermore, we also omit \mil{} because
the required memory for a batch size of 1 exceeds the available GPU memory.

\subsubsection{Performance}

Initially, we examine the effect of the number of sampled patches on the
performance of our method. We sample $\{5, 10, 25, 50, 100\}$ patches for each
image which corresponds to $0.01\%$ to $0.3\%$ of the available sampling space.
We train our models $5$ independent runs for $500$ epochs and the averaged
results are depicted in figures \ref{fig:megamnist_performance:a} and
\ref{fig:megamnist_performance:b}.  The figures show the training loss and test
error, respectively, with respect to wall clock time both for ATS and the CNN
baseline. Even though the CNN has comparably increased capacity, we observe
that ATS is order of magnitudes faster and performs better.

As expected, we observe that \ats{} directly trades performance for speed,
namely sampling fewer patches results in both higher training loss and test
error. Although the CNN baseline performs better than random guessing,
achieving roughly 40\% error, it is still more than an order of magnitude
higher than ATS. 

\subsubsection{Evolution of the attention distribution}

The quantitative results of the previous section demonstrate that \ats{}
processes high resolution images both faster and more accurately than the
CNN baseline. However, another important benefit of using attention is the
increased interpretability of the decisions of the network. This can be
noticed from Figure \ref{fig:megamnist_attention}, where we visualize the
evolution of the attention distribution as the training progresses. In
particular, we select a patch from a random image from the dataset that
contains 6 distinct items, 3 pieces of noise and 3 digits, and draw the
attention distribution for that patch. We observe that the attention
distribution starts as uniform.  However, during training, we note that the
attention network first learns to distinguish empty space from noise and digits
and subsequently even noise from digits. This explains why by only sampling 5
patches we achieve approximately $20\%$ error, even though it is the minimum
required to be able to confidently classify an image.

\subsection{Histopathology images} \label{sec:crch}

\begin{table*}[h]
    \centering
    \scalebox{0.8}{
    \begin{tabular}{lcccrr}
    \toprule
    Method & Scale & Train Loss & Test Error & Time/sample & Memory/sample \\
    \midrule
    U-10 & 0.2/1 & 0.210 $\pm$ 0.031 & 0.156 $\pm$ 0.006 & 1.8 ms  & 19 MB \\
    U-50 & 0.2/1 & 0.075 $\pm$ 0.000 & 0.124 $\pm$ 0.010 & 4.6 ms  & 24 MB \\
    CNN  & 0.5   & 0.002 $\pm$ 0.000 & 0.104 $\pm$ 0.009 & 4.8 ms  & 65 MB \\
    CNN  & 1     & 0.002 $\pm$ 0.000 & 0.092 $\pm$ 0.012 & 18.7 ms & 250 MB \\
    \mil{} \cite{ilse18a} & 1 & 0.007 $\pm$ 0.000 & 0.093 $\pm$ 0.004 & 48.5 ms   & 644 MB \\
    \midrule
    ATS-10 & 0.2/1 & 0.083 $\pm$ 0.019 & 0.093 $\pm$ 0.014 & 1.8 ms & 21 MB \\
    ATS-50 & 0.2/1 & 0.028 $\pm$ 0.002 & 0.093 $\pm$ 0.019 & 4.5 ms  & 26 MB \\
    \bottomrule
    \end{tabular}}
    \caption{Performance comparison of \ats{} (ATS) with a
             CNN and \mil{} on the \emph{colon cancer} dataset comprised of
             H\&E stained images. The experiments were run 5 times and the
             average ($\pm$ a standard error of the mean) is reported. ATS
             performs equally well to \mil{} and CNN in terms of test error,
             while being at least \textbf{10x} faster.}
    \label{tab:crch}
    \vspace{-1.2em}
\end{table*}

In this experiment, we evaluate \ats{} on the \emph{colon cancer} dataset
introduced by \citet{sirinukunwattana2016locality} to detect whether epithelial
cells exist in a hematoxylin and eosin (H\&E) stained image.

This dataset contains 100 images of dimensions $500 \times 500$. The images
originate both from malignant and normal tissue and contain approximately
22,000 annotated cells. Following the experimental setup of \citet{ilse18a},
we treat the problem as binary classification where the positive images are the
ones that contain at least one cell belonging in the epithelial class. While
the size of the images in this dataset is less than one
megapixel, our method can easily scale to datasets with much larger images, as
the computational and memory requirements depend only on the size and the number of
the patches. However, this does not apply to our baselines, where both the
memory and the computational requirements scale linearly with the size of the
input image. As a result, this experiment is a best case scenario for our
baselines.
\begin{figure}
    \centering
    \begin{subfigure}[b]{0.24\columnwidth}
        \includegraphics[width=\textwidth]{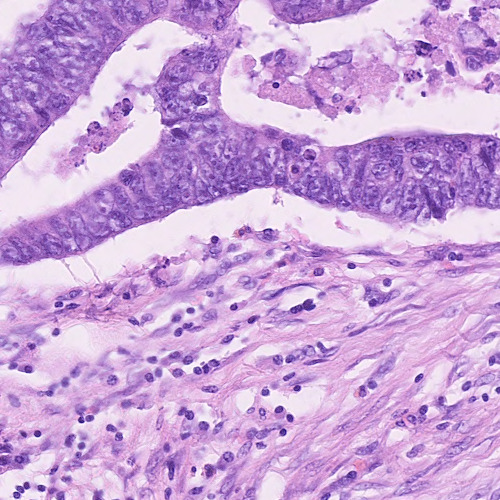}
        \caption{}
    \end{subfigure}
    \begin{subfigure}[b]{0.24\columnwidth}
        \includegraphics[width=\textwidth]{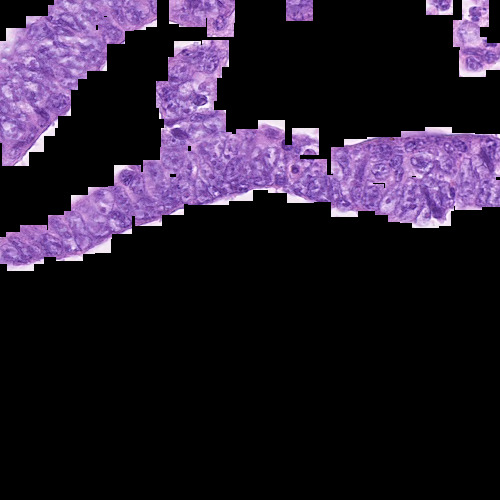}
        \caption{} \label{fig:crch_attention:gt}
    \end{subfigure}
    \begin{subfigure}[b]{0.24\columnwidth}
        \includegraphics[width=\textwidth]{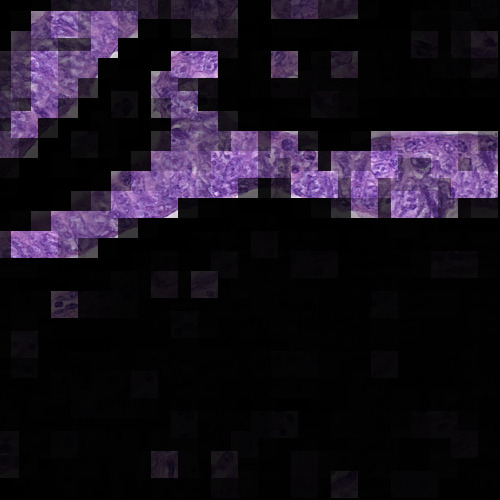}
        \caption{}
    \end{subfigure}
    \begin{subfigure}[b]{0.24\columnwidth}
        \includegraphics[width=\textwidth]{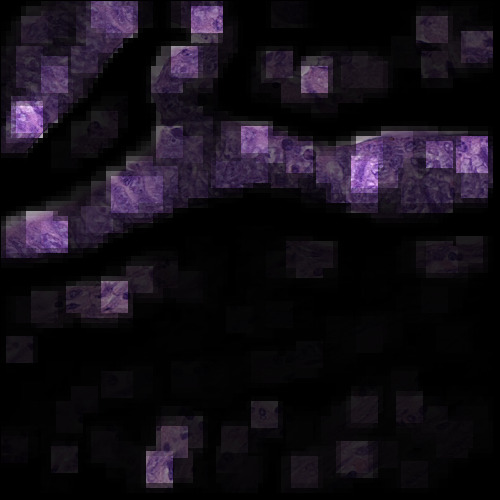}
        \caption{}
    \end{subfigure}
    \caption{Visualization of the learned attention distributions for \mil{}
             (c) and our \ats{} (d) on an H\&E stained image from the
             \emph{colon cancer} dataset. (a) depicts the raw image and (b)
             depicts the cells that belong to the epithelial class. Images (c)
             and (d) are created by multiplying every patch in (a) by the
             corresponding normalized attention weight. Both methods localize the attention
             distribution effectively on the informative parts of the
             image.}
    \label{fig:crch_attention}
    \vspace{-1.2em}
\end{figure}

For our model, we downsample the images by a factor of 5 and we use the
attention network described in \S~\ref{sec:megamnist}.  The feature network of
our model is the same as the one proposed by \citet{ilse18a} with input patches
of size $27 \times 27$. For \mil{}, we extract 2,500 patches per image at a
regular grid. Regarding the CNN baseline, we use a ResNet \cite{he2016deep}
architecture. Furthermore, we perform data
augmentation by small random adjustments to the brightness and contrast of each
image. Following \citet{ilse18a}, we perform 5 independent runs and report the
mean and the standard error of the mean.

\subsubsection{Performance}

The results of this experiment are summarized in Table~\ref{tab:crch}. We
observe that sampling from the uniform distribution 10 and 50
patches is clearly better than random guessing by achieving 15.6\% and 12.4\%
error respectively. This stems from the fact that each positive sample
contains hundreds of regions of interest, namely epithelial cells, and we only
need one to classify the image.  As expected, \ats{}
learns to focus only on informative parts of the image thus resulting in
approximately 35\% lower test error and 3 times lower training loss.
Furthermore, compared to \mil{} and CNN, ATS-10 performs equally well while
being \textbf{25x} and \textbf{10x} faster respectively. Moreover, the most
memory efficient baseline (CNN) needs at least \textbf{3x} more memory compared
to \ats{}, while \mil{} needs \textbf{30x} more.

\subsubsection{Attention Distribution}

To show that our proposed model indeed learns to focus on informative
parts of the image, we visualize the learned attention distribution at the end
of training. In particular, we select an image from the test set and we compute
the attention distribution both for \mil{} and \ats{}. Subsequently, we weigh
each corresponding patch with a normalized attention value that is computed as
$w_i = \frac{a_i - \min(a)}{\max(a) - \min(a)}$. For reference, in
Figure~\ref{fig:crch_attention}, apart from the two attention distributions, we
also visualize the patches that contain an epithelial cell.
Both models identify epithelial
cells without having access to per-patch annotations.
In order to properly classify an image as positive or not, we just need
to find a single patch that contains an epithelial cell. Therefore, despite the
fact that the learned attention using \ats{} matches less well the distribution
of the epithelial cells (Figure~\ref{fig:crch_attention:gt}), compared to
\mil{}, it is not necessarily worse for the classification task that we are
interested in. However, it is less helpful for detecting regions of interest.
In addition, we also observe that both attentions have significant overlap even
on mistakenly selected patches such as the bottom center of the images.

\subsection{Speed limit sign detection} \label{sec:speedlimits}

\begin{table*}[h]
    \centering
    \scalebox{0.8}{
    \begin{tabular}{lcccrr}
    \toprule
    Method & Scale & Train Loss & Test Error & Time/sample & Memory/sample \\
    \midrule
    U-5    & 0.3/1 & 1.468 $\pm$ 0.317 & 0.531 $\pm$ 0.004 & 7.8 ms  & 39 MB \\
    U-10   & 0.3/1 & 0.851 $\pm$ 0.408 & 0.472 $\pm$ 0.008 & 10.8 ms  & 78 MB \\
    CNN    & 0.3   & 0.003 $\pm$ 0.001 & 0.311 $\pm$ 0.049 & 6.6 ms  & 86 MB \\
    CNN    & 0.5   & 0.002 $\pm$ 0.001 & 0.295 $\pm$ 0.039 & 15.6 ms & 239 MB \\
    CNN    & 1     & 0.002 $\pm$ 0.000 & 0.247 $\pm$ 0.001 & 64.2 ms  & 958 MB \\
    \mil{} \cite{ilse18a} & 1     & 0.077 $\pm$ 0.089 & 0.083 $\pm$ 0.006 & 97.2 ms  & 1,497 MB \\
    \midrule
    ATS-5  & 0.3/1 & 0.162 $\pm$ 0.124 & 0.089 $\pm$ 0.002 & 8.5 ms  & 86 MB \\
    ATS-10 & 0.3/1 & 0.082 $\pm$ 0.032 & 0.095 $\pm$ 0.008 & 10.3 ms & 118 MB \\
    \bottomrule
    \end{tabular}}
    \caption{Performance comparison of \ats{} (ATS) with a
             CNN and \mil{} on the \emph{speed limits} dataset. The experiments
             were run 3 times and the average ($\pm$ a standard error of the
             mean) is reported.
             ATS performs equally well as \mil{} while being at least
             \textbf{10x} faster.  Regarding CNN, we note that both \mil{} and
             \ats{} perform significantly better.}
    \label{tab:speedlimits}
    \vspace{-1.2em}
\end{table*}

\begin{figure}
    \centering
    \begin{subfigure}[b]{0.25\columnwidth}
        \includegraphics[width=\textwidth]{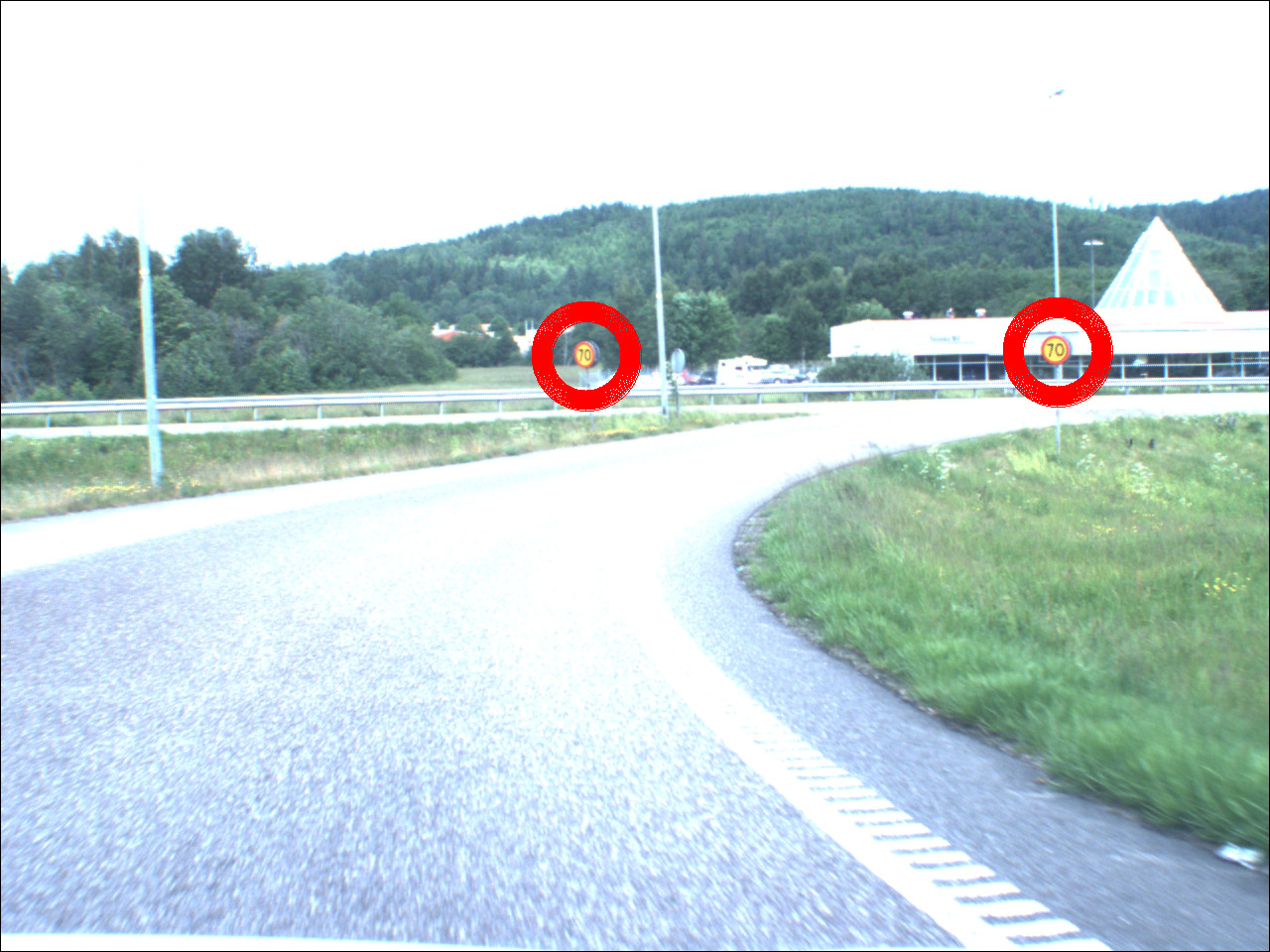}
        \caption{} \label{fig:speed_limits_attention:gt}
        \vspace{1.5em}
    \end{subfigure}
    \hfill
    \begin{subfigure}[b]{0.25\columnwidth}
        \includegraphics[width=\textwidth]{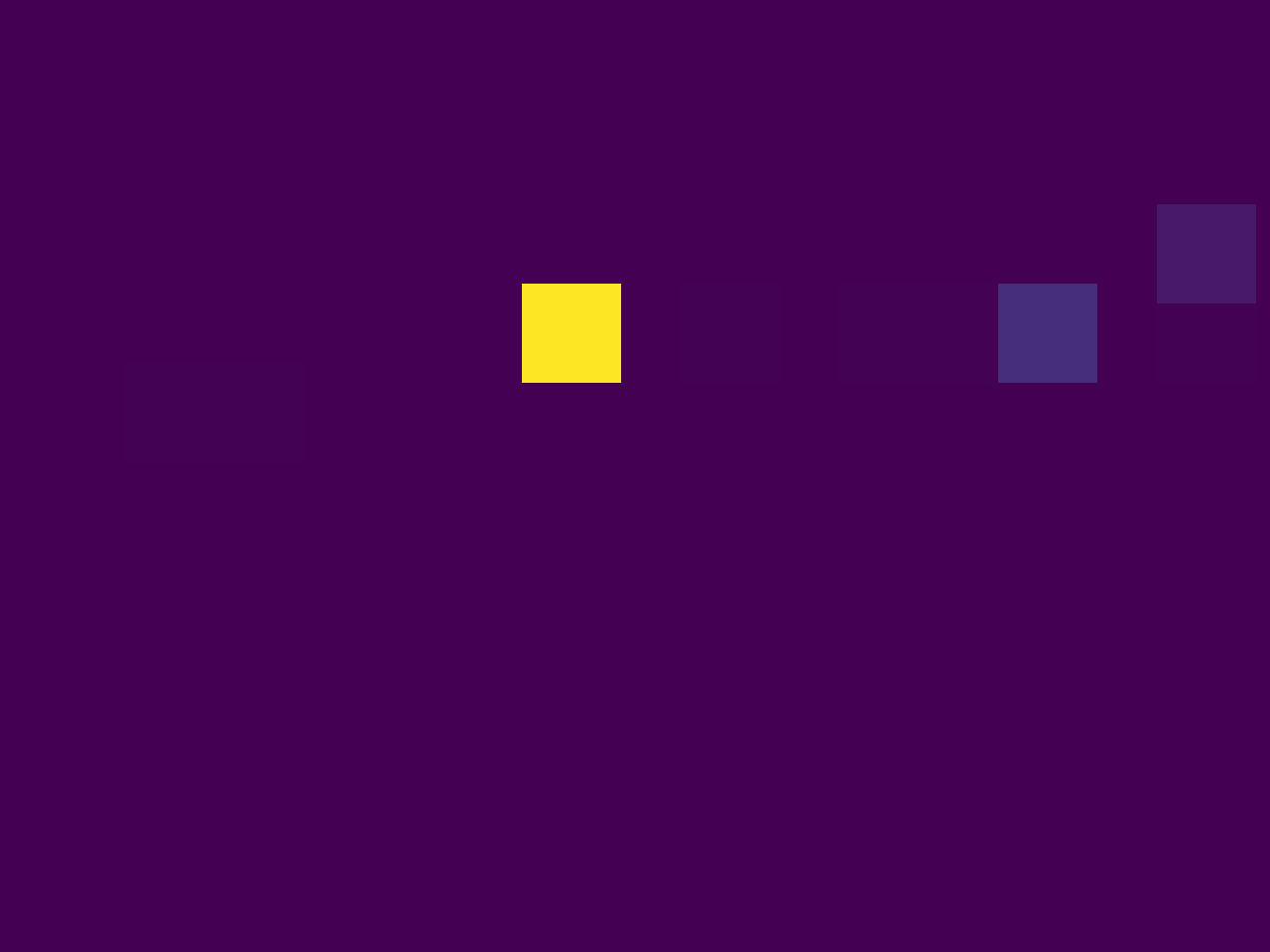}
        \caption{}
        \vspace{1.5em}
    \end{subfigure}
    \hfill
    \begin{subfigure}[b]{0.25\columnwidth}
        \includegraphics[width=\textwidth]{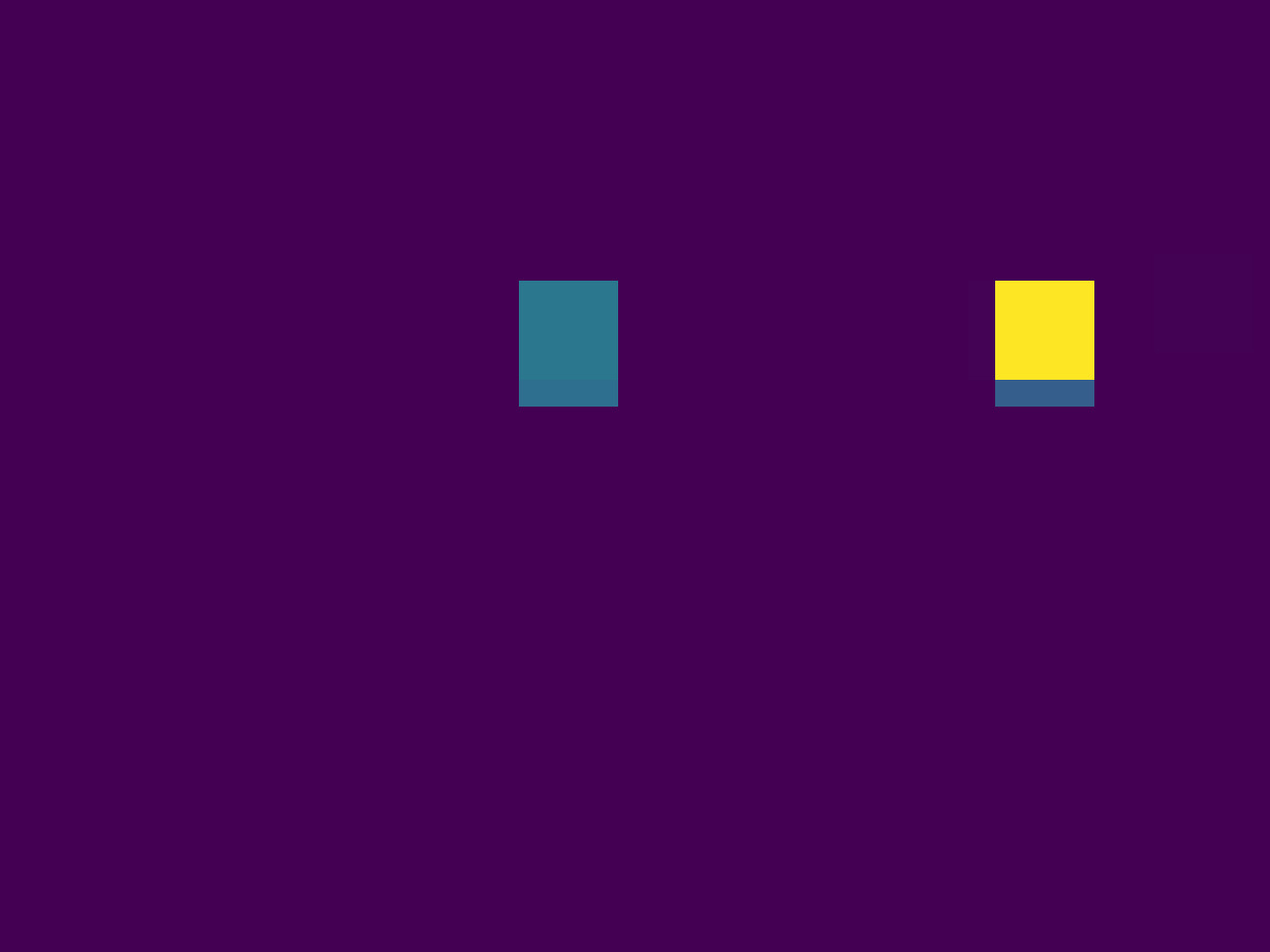}
        \caption{}
        \vspace{1.5em}
    \end{subfigure}
    \hfill
    \begin{minipage}[b]{0.11\columnwidth}
        \begin{subfigure}[b]{\textwidth}
            \centering
            \includegraphics[width=\textwidth]{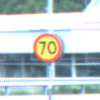}
            \caption{} \label{fig:speed_limits_attention:p1}
        \end{subfigure}
        \begin{subfigure}[b]{\textwidth}
            \centering
            \includegraphics[width=\textwidth]{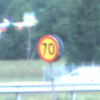}
            \caption{} \label{fig:speed_limits_attention:p2}
        \end{subfigure}
    \end{minipage}
    \caption{Visualization of the learned attention distributions for \mil{}
             (b) and our \ats{} (c) on an image from the speed limits dataset
             (a). (d) and (e) depict the marked regions from (a) which are also
             selected by \ats{}. We observe that both of them contain speed
             limit signs unrecognizable in the low resolution image.}
    \label{fig:speed_limits_attention}
    \vspace{-1.2em}
\end{figure}

In this experiment, we seek to classify images based on whether they contain no
speed limit or a limit sign of $50$, $70$ or $80$ kilometers per hour. We use a
subset of the Swedish traffic signs dataset \cite{larsson2011using}, for which
we do not use explicit annotations of the signs, just one label for each
image.
The dataset contains $3,777$ images annotated
with 20 different traffic sign classes. Each image is 1.3 megapixels, namely
$960 \times 1280$ pixels. As some classes contain less than $20$
samples, we limit the classification task to the one described above. The
resulting dataset consists of 747 training images and 684 test images,
distributed approximately as 100 images for each speed limit sign and 400 for
the background class, namely no limit sign.

An interesting fact about this dataset is that in order to properly classify
all images it is mandatory to process them in high resolution. This is
illustrated in Figure~\ref{fig:sign_far}, where from the downsampled image one
can deduce the existence of a speed limit sign, without being able
to identify the number of kilometers written on it. Objects that are physically
far from the moving camera become unrecognizable when downsampling the input
image. This property might be critical, for early detection of pedestrians or
collision avoidance in a self-driving car scenario.

For \ats{}, we downsample the original image by approximately a factor of $3$
to $288\times384$. The attention network is a four layer convolutional network
and the feature network of both our model and \mil{} is a simple ResNet. For
\mil{}, we extract $192$ patches on a grid $12 \times 16$ of patch size
$100\times 100$.  For a fair comparison, we evaluate the CNN baseline using
images at various resolutions, namely scales $0.3$, $0.5$ and $1.0$.

Again also for this dataset, we perform data augmentation, namely random
translations and contrast brightness adjustments. In addition, due to class
imbalance, for all evaluated methods, we use a crossentropy loss weighted with
the inverse of the prior of each class. We perform 3 independent runs and
report the mean and the standard error of the mean.

\subsubsection{Performance}

Table \ref{tab:speedlimits} compares the proposed model to our baselines on the
speed limits dataset. We observe that although the CNN learns the
training set perfectly, it fails to generalise. For the downsampled images, this
is expected as the limits on the traffic signs are indistinguishable.
Similarly, due to the small number of informative patches, uniform
sampling fails to correctly classify both the training set and the test set.
We observe that \ats{} achieves comparable test error to \mil{} by using
just $5$ patches, instead of $192$. This results in significant speedups of
more than an order of magnitude. Regarding the required memory, \ats{} needs
\textbf{17x} less memory compared to \mil{}.

\subsubsection{Attention Distribution}

In this section, we compare qualitatively the learned attention distribution of
\mil{} and \ats{} on an image from the test set of the speed limits dataset. In
Figure~\ref{fig:speed_limits_attention:gt}, we mark the positions of speed
limit signs with red circles and visualize the corresponding patches in figures
\ref{fig:speed_limits_attention:p1} and \ref{fig:speed_limits_attention:p2}. We
observe that the attention distribution from our proposed model has high
probability for both patches whereas \mil{} locates both but selects only one.
Also in this dataset, both models identify regions of interest in the images
without being given any explicit per-patch
label.

\section{Conclusions}

We have presented a novel algorithm to efficiently process megapixel images in
a single CPU or GPU. Our algorithm only processes fractions of the input image,
relying on an attention distribution to discover informative regions of the
input. We show that we can derive the gradients through the sampling and train
our model end-to-end with SGD. Furthermore, we show that sampling with the
attention distribution is the optimal approximation, in terms of variance, of
the model that processes the whole image.

Our experiments show that our algorithm effectively identifies the important
regions in two real world tasks and an artificial dataset without any patch
specific annotation. In addition, our model executes an order of magnitude
faster and requires an order of magnitude less memory than state of the art
patch based methods and traditional CNNs.

The presented line of research opens several directions for future work. We
believe that a nested model of \ats{} can be used to efficiently learn to
discover informative regions and classify up to gigapixel images using a single
GPU. In addition, \ats{} can be used in resource constrained scenarios to
finely control the trade-off between accuracy and spent computation.

\section*{Acknowledgement}

This work is supported by the Swiss National Science Foundation under grant
number FNS-30209 ``ISUL''.

\bibliographystyle{icml2019}
\bibliography{references}

\appendix
\renewcommand*{\thesubfigure}{\arabic{subfigure}}

\clearpage
\twocolumn[
    \standalonetitle{Appendix}
    \vspace{4em}
]

\section{Introduction}

This supplementary material is organised as follows: In \S~\ref{sec:grad} and
\S~\ref{sec:grad_swor} we provide the detailed derivation of the gradients for
our \ats{}. Subsequently, in \S~\ref{sec:supp_related} we mention additional
related work that might be of interest to the readers. In
\S~\ref{sec:entropy_ablation} and \S~\ref{sec:patches_ablation} we present
experiments that analyse the effect of our entropy regularizer and the number
of patches sampled on the learned attention distribution. In
\S~\ref{sec:qualitative}, we visualize the attention distribution of our method
to show it focuses computation on the informative parts of the high resolution
images. Finally, in \S~\ref{sec:networks} we provide details with respect to
the architectures trained for our experiments.

\section{Sampling with replacement} \label{sec:grad}
In this section, we detail the derivation of equation $11$ in our main
submission. In order to be able to use a neural network as our attention
distribution we need to derive the gradient of the loss with respect to the
parameters of the attention
function $a(\cdot; \Theta)$ through the sampling of the set of indices $Q$.
Namely, we need to compute
\begin{align} \label{eq:supp_grad_initial}
\diff{\frac{1}{N} \sum_{q \in Q} f(x; \Theta)_q}{\theta}
\end{align}
for all $\theta \in \Theta$ including the ones that affect $a(\cdot)$.

By exploiting the Monte Carlo approximation and the multiply by one trick, we
get
\begin{align}
& \diff{}{\theta}\frac{1}{N} \sum_{q \in Q} f(x; \Theta)_q \label{eq:grad_mc} \\
&\quad \approx \diff{}{\theta}\sum_{i=1}^K a(x; \Theta)_i f(x; \Theta)_i
    \label{eq:grad_mc_2} \\
&\quad = \sum_{i=1}^K \diff{}{\theta}\left[a(x; \Theta)_i f(x; \Theta)_i\right] 
    \label{eq:grad_mc_3} \\
&\quad = \sum_{i=1}^K \frac{a(x; \Theta)_i}{a(x; \Theta)_i}
    \diff{}{\theta}\left[a(x; \Theta)_i f(x; \Theta)_i\right]  \label{eq:grad_mc_4} \\
&\quad = \E[I \sim a(x; \Theta)]{
    \frac{\diff{}{\theta}\left[a(x; \Theta)_I f(x; \Theta)_I\right]}
        {a(x; \Theta)_I}
    }. \label{eq:supp_grad_final}
\end{align}

\section{Sampling without replacement} \label{sec:grad_swor}

In this section, we derive the gradients of the attention distribution with
respect to the feature network and attention network parameters. We define
\begin{itemize}
    \item $f_i=f(x;\Theta)_i$ for $i \in \{1, 2, \dots, K\}$ to be the $K$ features
    \item $a_i=a(x;\Theta)_i$ for $i \in \{1, 2, \dots, K\}$ to be the
        probability of the $i$-th feature from the attention distribution $a$
    \item $w_i = \sum_{j \ne i} a_j$
\end{itemize}

We consider sampling without replacement to be sampling an index $i$ from $a$
and then sampling from the distribution $p_i(j)$ defined for $j \in \{1,
2,\dots,i-1,i+1,\dots,K\}$ as follows,
\begin{align}
    p_i(j) = \frac{a_j}{w_i}.
\end{align}

Given samples $i, j$ sampled from $a$ and $p_i$, we can make an unbiased
estimator for $\E[I \sim a]{f_I}$ as follows,
\begin{align}
    a_i f_i + w_i f_j &\simeq \\
    \E[I \sim a]{\E[J \sim p_I]{a_I f_I + w_I f_J}} &= \\
    \E[I \sim a]{a_I f_I + \E[J \sim p_I]{w_I f_J}} &= \\
    \E[I \sim a]{a_I f_I + \sum_{j \ne I} a_j f_j} &= \\
    \E[I \sim a]{\sum_{j=1}^K a_j f_j} &= \\
    \E[I \sim a]{f_I}.
\end{align}

Using the same $i, j$ sampled from $a$ and $p_i$ accordingly, we can estimate
the gradient as follows,
\begin{align}
    \diff{}{\theta}\E[I \sim a]{f_I} &= \\
    \diff{}{\theta}\E[I \sim a]{\E[J \sim p_I]{a_I f_I + w_I f_J}} &= \\
    \diff{}{\theta} \sum_{i=1}^K \sum_{j \ne i}
        a_i p_i(j) \left(a_i f_i + w_i f_j\right) &= \\
    \sum_{i=1}^K \sum_{j \ne i} \diff{}{\theta} 
        a_i p_i(j) \left(a_i f_i + w_i f_j\right) &= \\
    \sum_{i=1}^K \sum_{j \ne i} \frac{a_i p_i(j)}{a_i p_i(j)}
        \diff{}{\theta} a_i p_i(j) \left(a_i f_i + w_I f_j\right) &= \\
    \E[I \sim a]{\E[J \sim p_I]{
        \frac{\diff{}{\theta} a_I p_I(J) \left(a_I f_I + w_I f_J\right)}
            {a_I p_I(J)}
        }} &\simeq \\
    \frac{\diff{}{\theta} a_i p_i(j) \left(a_i f_i + w_i f_j\right)}
        {a_i p_i(j)} &= \label{eq:supp_mc_grad} \\
    \frac{p_i(j) \left(a_i f_i + w_i f_j\right)\diff{}{\theta} a_i}
        {a_i p_i(j)} +
        \frac{a_i \diff{}{\theta}p_i(j) \left(a_i f_i + w_i f_j\right)}
        {a_i p_i(j)} &= \\
    \left(a_i f_i + w_i f_j\right) \frac{\diff{}{\theta} a_i}{a_i} +
        \frac{\diff{}{\theta}p_i(j) \left(a_i f_i + w_i f_j\right)}
        {p_i(j)} &= \\
    \left(a_i f_i + w_i f_j\right) \diff{}{\theta}\log(a_i) +
        \frac{\diff{}{\theta}p_i(j) \left(a_i f_i + w_i f_j\right)}
        {p_i(j)}. \label{eq:supp_final_grad}
\end{align}

When we extend the above computations for sampling more than two samples, the
logarithm in equation \ref{eq:supp_final_grad} allows us to avoid the numerical
errors that arise from the cumulative product at equation \ref{eq:supp_mc_grad}.

\section{Extra related work} \label{sec:supp_related}

For completeness, in this section we discuss parts of the literature that are
tangentially related to our work.

\citet{combalia2018monte} consider the problem of high-resolution image
classification from the Multiple Instance Learning perspective. The authors
propose a two-step procedure; initially random patches are sampled and
classified. Subsequently, more patches are sampled around the patches that
resulted in confident predictions. The most confident prediction is returned.
Due to the lack of the attention mechanism, this model relies in identifying
the region of interest via the initial random patches. However, in the second
pass the prediction is finetuned if informative patches are likely to be
spatially close with each other.

\citet{maggiori2017high} propose a neural network architecture for the
pixelwise classification of high resolution images. The authors consider
features at several resolutions and train a pixel-by-pixel fully connected
network to combine the features into the final classification. The
aforementioned approach could be used with our \ats{} to approach pixelwise
classification tasks such as semantic segmentation.

\section{Ablation study on the entropy regularizer} \label{sec:entropy_ablation}

\begin{figure*}
    \captionsetup[subfigure]{labelformat=empty}
    \centering
    \begin{subfigure}[b]{0.2\textwidth}
        \includegraphics[width=\textwidth]{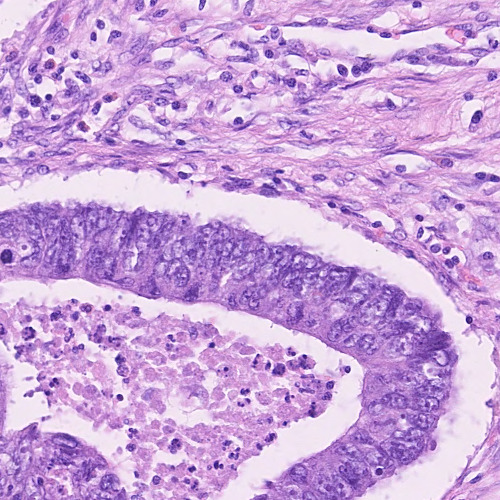}
        \caption{Sample image}
        \vspace{10em}
    \end{subfigure}
    \hfill
    \begin{minipage}[b]{0.1\textwidth}
        \begin{subfigure}[b]{\textwidth}
            \hfill $\lambda=0$
            \vspace{6.5em}
        \end{subfigure}
        \begin{subfigure}[b]{\textwidth}
            \hfill $\lambda=0.01$
            \vspace{6.5em}
        \end{subfigure}
        \begin{subfigure}[b]{\textwidth}
            \hfill $\lambda=0.1$
            \vspace{6.5em}
        \end{subfigure}
        \begin{subfigure}[b]{\textwidth}
            \hfill $\lambda=1$
            \vspace{4.8em}
        \end{subfigure}
    \end{minipage}
    \begin{minipage}[b]{0.63\textwidth}
        \begin{subfigure}[b]{0.24\textwidth}
            \includegraphics[width=\textwidth]{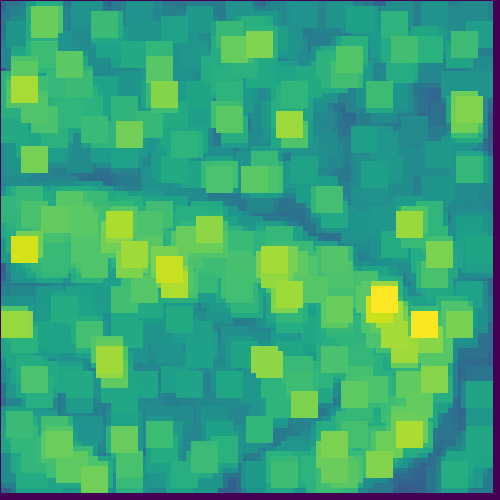}
        \end{subfigure}
        \begin{subfigure}[b]{0.24\textwidth}
            \includegraphics[width=\textwidth]{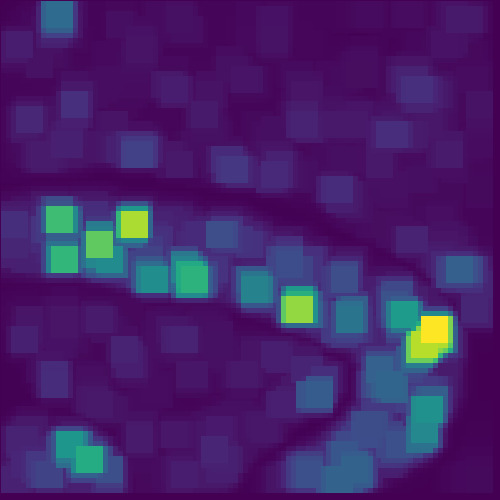}
        \end{subfigure}
        \begin{subfigure}[b]{0.24\textwidth}
            \includegraphics[width=\textwidth]{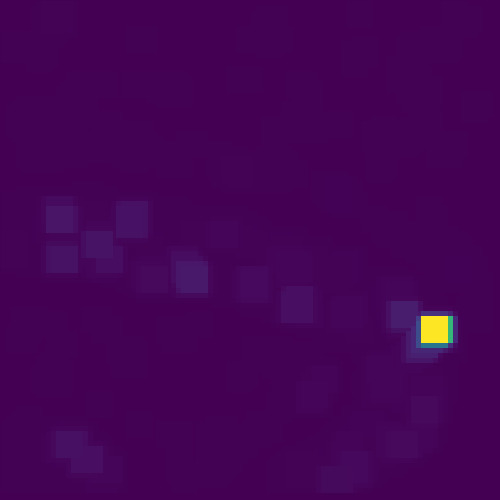}
        \end{subfigure}
        \begin{subfigure}[b]{0.24\textwidth}
            \includegraphics[width=\textwidth]{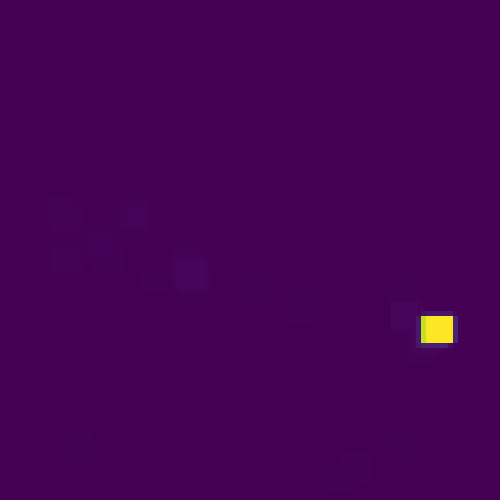}
        \end{subfigure}

        \begin{subfigure}[b]{0.24\textwidth}
            \includegraphics[width=\textwidth]{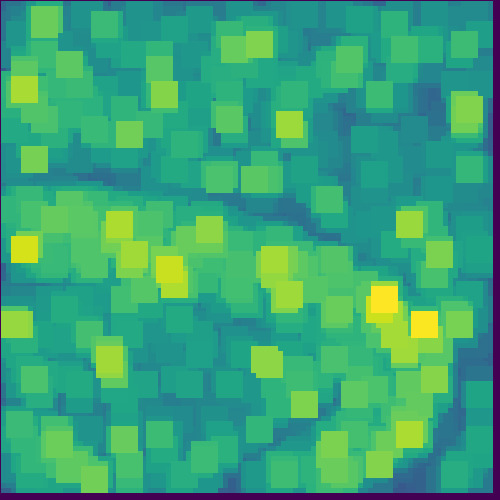}
        \end{subfigure}
        \begin{subfigure}[b]{0.24\textwidth}
            \includegraphics[width=\textwidth]{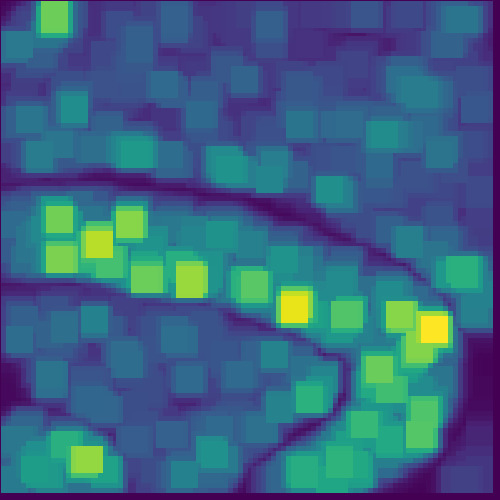}
        \end{subfigure}
        \begin{subfigure}[b]{0.24\textwidth}
            \includegraphics[width=\textwidth]{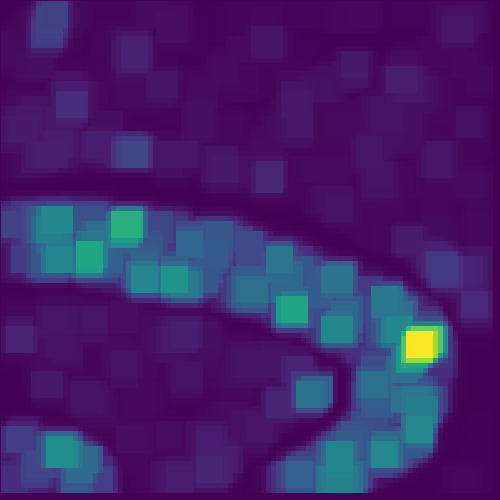}
        \end{subfigure}
        \begin{subfigure}[b]{0.24\textwidth}
            \includegraphics[width=\textwidth]{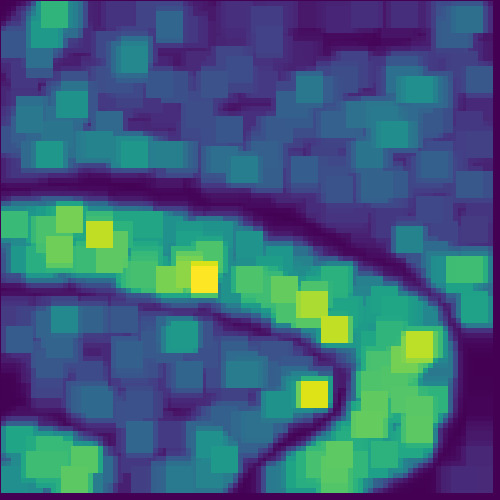}
        \end{subfigure}

        \begin{subfigure}[b]{0.24\textwidth}
            \includegraphics[width=\textwidth]{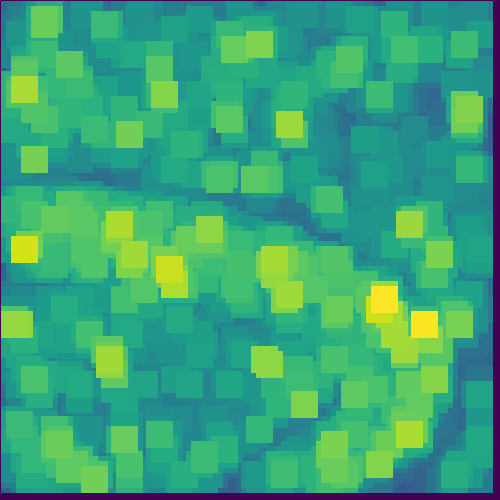}
        \end{subfigure}
        \begin{subfigure}[b]{0.24\textwidth}
            \includegraphics[width=\textwidth]{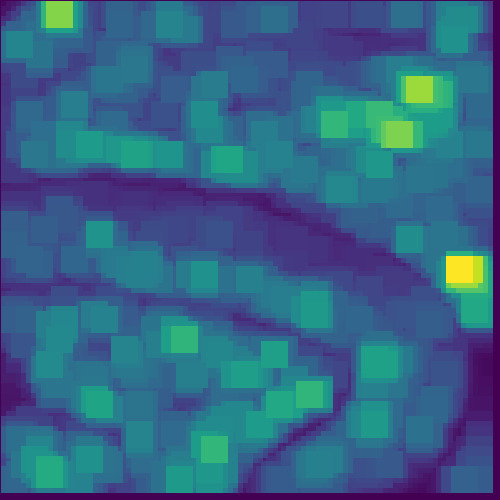}
        \end{subfigure}
        \begin{subfigure}[b]{0.24\textwidth}
            \includegraphics[width=\textwidth]{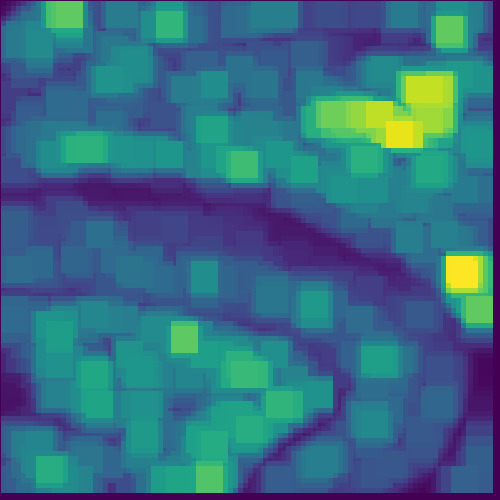}
        \end{subfigure}
        \begin{subfigure}[b]{0.24\textwidth}
            \includegraphics[width=\textwidth]{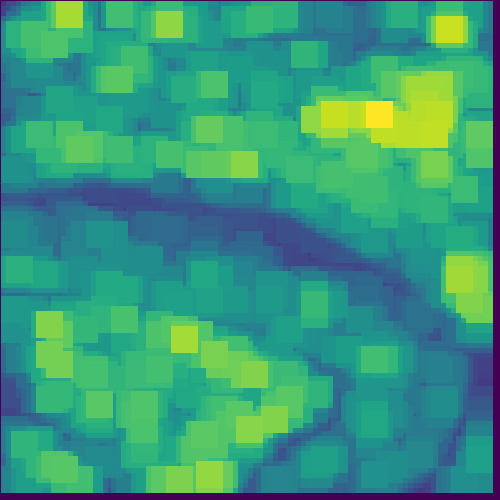}
        \end{subfigure}

        \begin{subfigure}[b]{0.24\textwidth}
            \includegraphics[width=\textwidth]{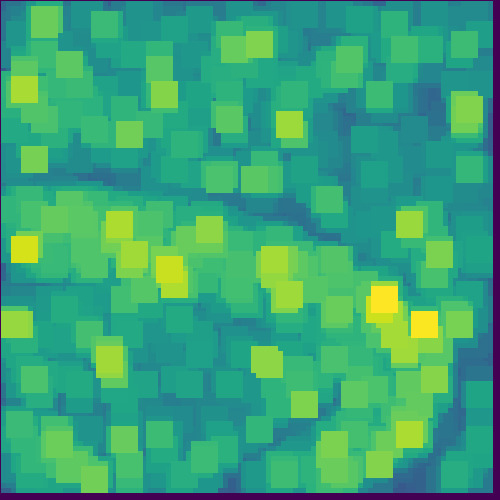}
            \caption{Epoch 0}
        \end{subfigure}
        \begin{subfigure}[b]{0.24\textwidth}
            \includegraphics[width=\textwidth]{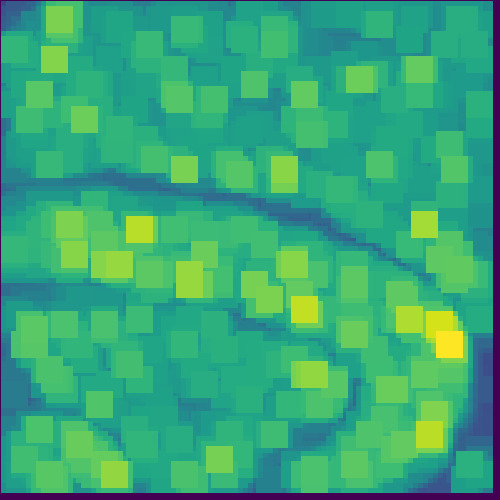}
            \caption{Epoch 20}
        \end{subfigure}
        \begin{subfigure}[b]{0.24\textwidth}
            \includegraphics[width=\textwidth]{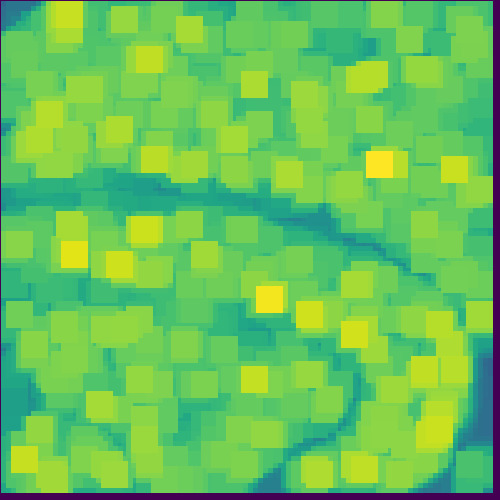}
            \caption{Epoch 40}
        \end{subfigure}
        \begin{subfigure}[b]{0.24\textwidth}
            \includegraphics[width=\textwidth]{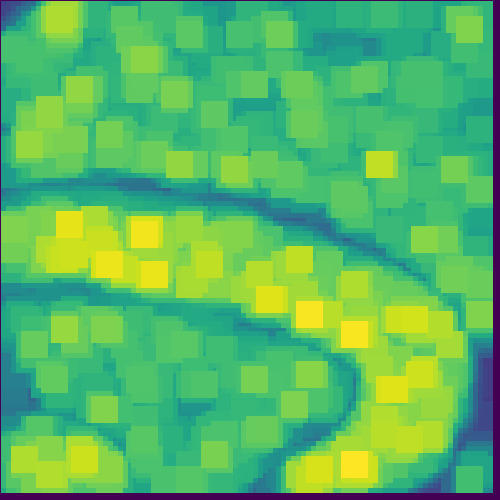}
            \caption{Epoch 60}
        \end{subfigure}
    \end{minipage}
    \caption{We visualize the effects of the entropy regularizer on the
             sampling distribution computeed from a test image of the
             \emph{colon cancer} dataset in the first 60 epochs of training.
             We observe that no entropy regularizer results in our attention
             becoming very selective early during training which might hinder
             the exploration of the sampling space.}
    \label{fig:entropy_ablation}
\end{figure*}

To characterize the effect of the entropy regularizer on our \ats{}, we train
with the same experimental setup as for the histopathology images of \S~4.3 but
varying the entropy regularizer $\lambda \in \{0, 0.01, 0.1, 1\}$. The results
are depicted in Figure~\ref{fig:entropy_ablation}. Using no entropy regularizer
results in a very selective attention distribution in the first 60 epochs of
training. On the other hand, a high value for $\lambda$, the entropy
regularizer weight, drives the sampling distribution towards uniform.

In our experiments we observed that values close to 0.01 (e.g.\ 0.005 or 0.05)
had no observable difference in terms of the final attention distribution.

\section{Ablation study on the number of patches} \label{sec:patches_ablation}

According to our theory, the number of patches should not affect the learned
attention distribution. Namely, the expectation of the gradients and the
predictions should be the same and the only difference is in the variance.

\begin{figure*}
    \captionsetup[subfigure]{labelformat=empty}
    \centering
    \begin{subfigure}[b]{0.2\textwidth}
        \includegraphics[width=\textwidth]{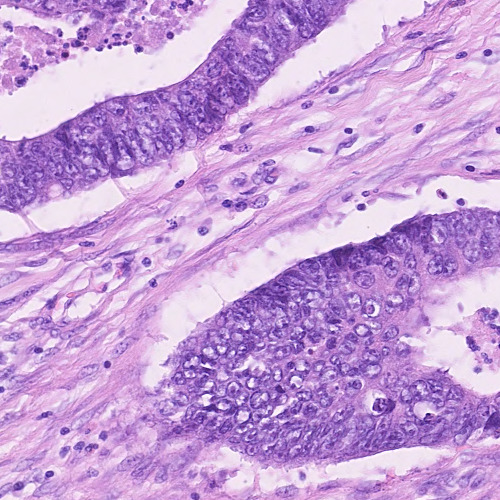}
        \caption{Sample image}
        \vspace{10em}
    \end{subfigure}
    \hfill
    \begin{minipage}[b]{0.1\textwidth}
        \begin{subfigure}[b]{\textwidth}
            \hfill 1 patch
            \vspace{6.5em}
        \end{subfigure}
        \begin{subfigure}[b]{\textwidth}
            \hfill 3 patches
            \vspace{6.5em}
        \end{subfigure}
        \begin{subfigure}[b]{\textwidth}
            \hfill 6 patches
            \vspace{6.5em}
        \end{subfigure}
        \begin{subfigure}[b]{\textwidth}
            \hfill 12 patches
            \vspace{4.8em}
        \end{subfigure}
    \end{minipage}
    \begin{minipage}[b]{0.63\textwidth}
        \begin{subfigure}[b]{0.24\textwidth}
            \includegraphics[width=\textwidth]{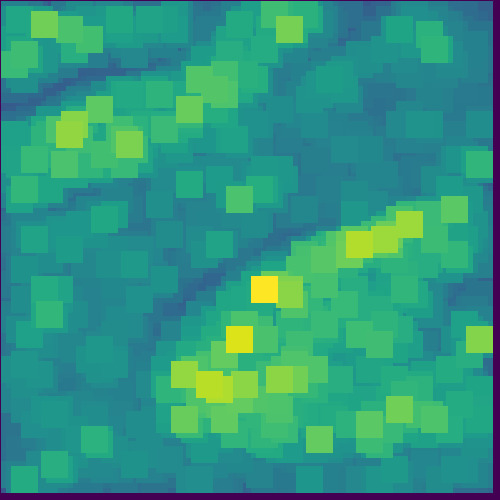}
        \end{subfigure}
        \begin{subfigure}[b]{0.24\textwidth}
            \includegraphics[width=\textwidth]{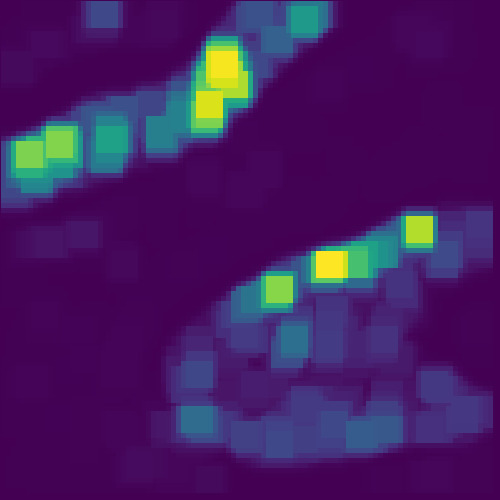}
        \end{subfigure}
        \begin{subfigure}[b]{0.24\textwidth}
            \includegraphics[width=\textwidth]{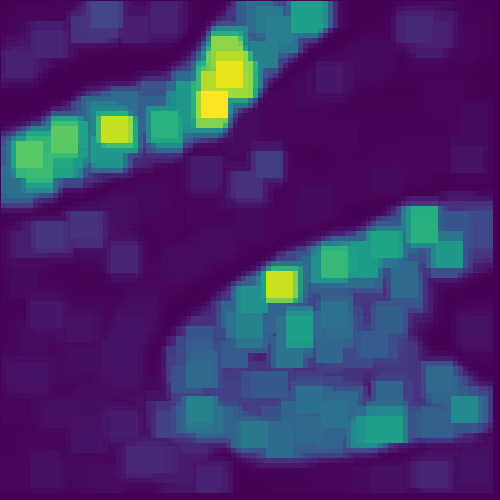}
        \end{subfigure}
        \begin{subfigure}[b]{0.24\textwidth}
            \includegraphics[width=\textwidth]{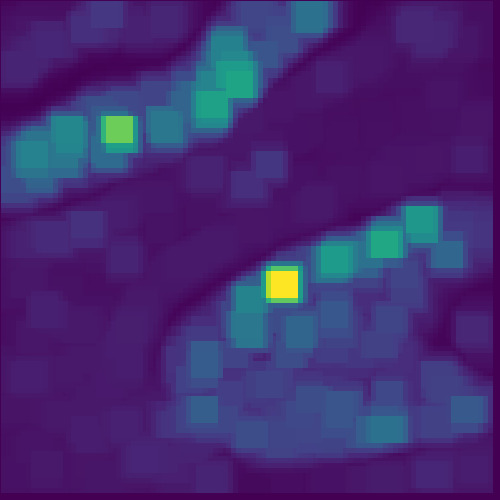}
        \end{subfigure}

        \begin{subfigure}[b]{0.24\textwidth}
            \includegraphics[width=\textwidth]{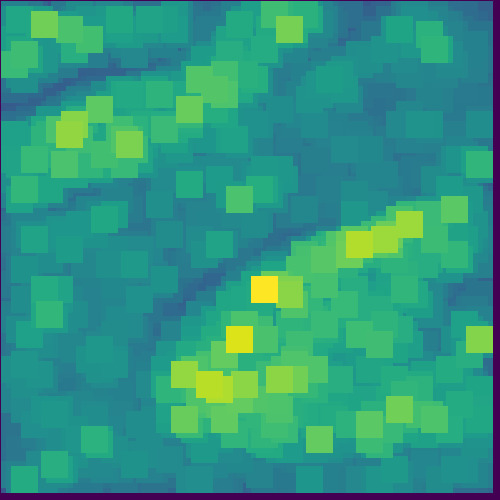}
        \end{subfigure}
        \begin{subfigure}[b]{0.24\textwidth}
            \includegraphics[width=\textwidth]{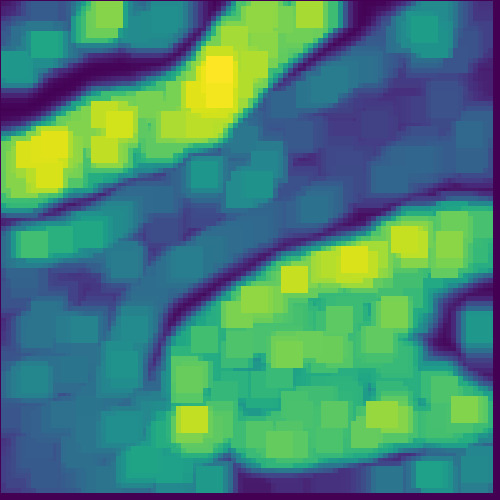}
        \end{subfigure}
        \begin{subfigure}[b]{0.24\textwidth}
            \includegraphics[width=\textwidth]{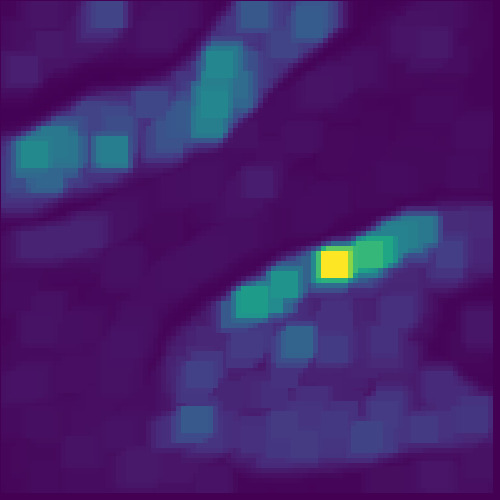}
        \end{subfigure}
        \begin{subfigure}[b]{0.24\textwidth}
            \includegraphics[width=\textwidth]{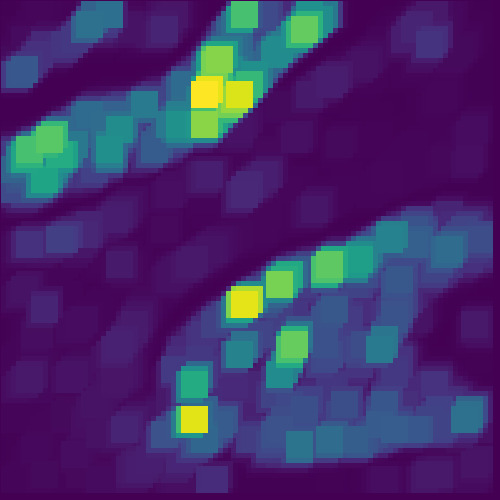}
        \end{subfigure}

        \begin{subfigure}[b]{0.24\textwidth}
            \includegraphics[width=\textwidth]{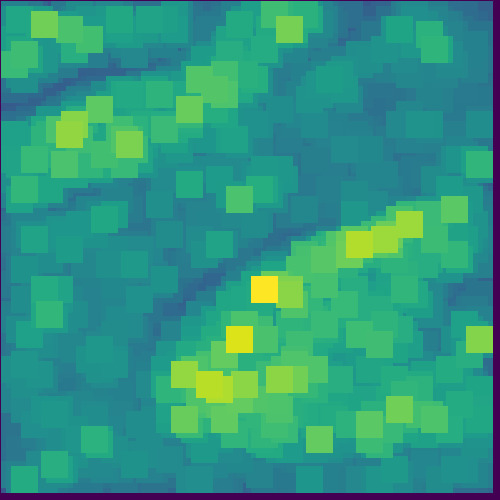}
        \end{subfigure}
        \begin{subfigure}[b]{0.24\textwidth}
            \includegraphics[width=\textwidth]{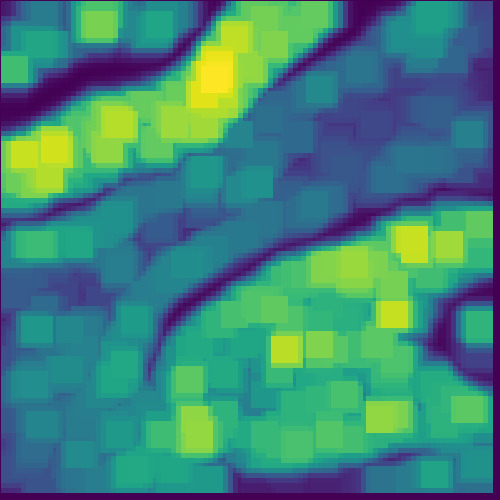}
        \end{subfigure}
        \begin{subfigure}[b]{0.24\textwidth}
            \includegraphics[width=\textwidth]{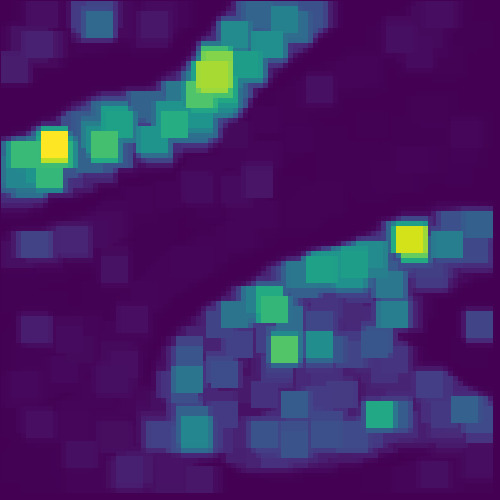}
        \end{subfigure}
        \begin{subfigure}[b]{0.24\textwidth}
            \includegraphics[width=\textwidth]{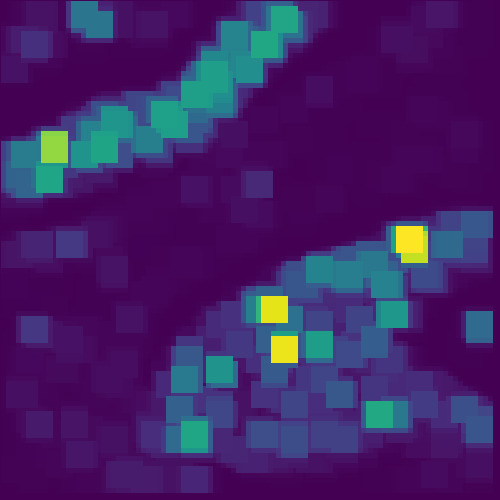}
        \end{subfigure}

        \begin{subfigure}[b]{0.24\textwidth}
            \includegraphics[width=\textwidth]{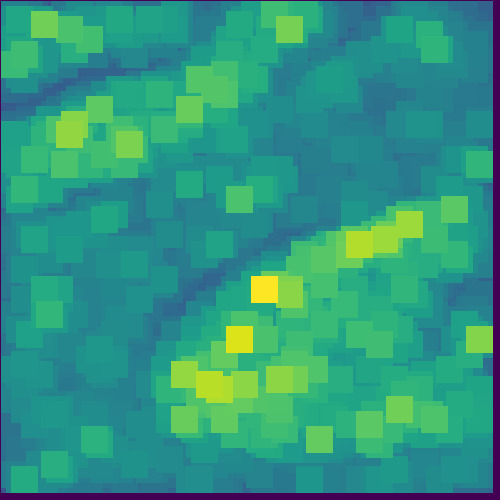}
            \caption{Epoch 0}
        \end{subfigure}
        \begin{subfigure}[b]{0.24\textwidth}
            \includegraphics[width=\textwidth]{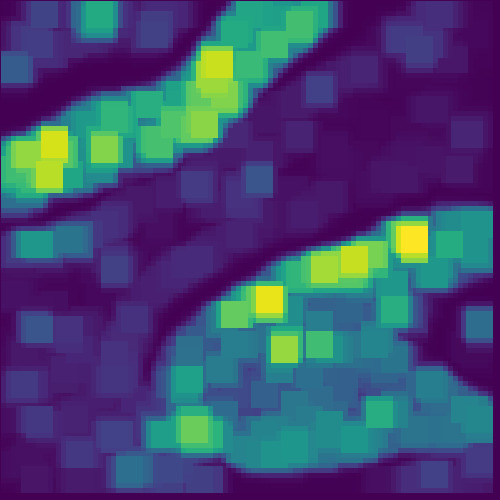}
            \caption{Epoch 40}
        \end{subfigure}
        \begin{subfigure}[b]{0.24\textwidth}
            \includegraphics[width=\textwidth]{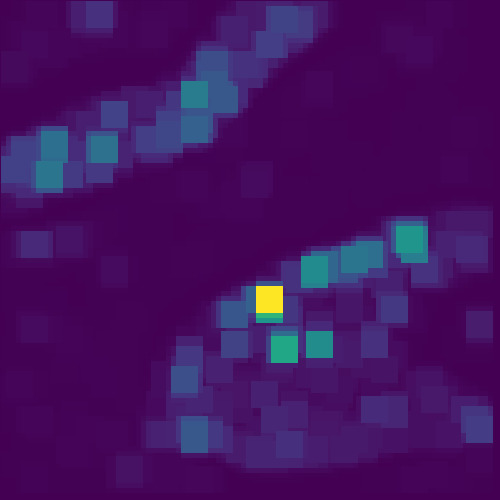}
            \caption{Epoch 80}
        \end{subfigure}
        \begin{subfigure}[b]{0.24\textwidth}
            \includegraphics[width=\textwidth]{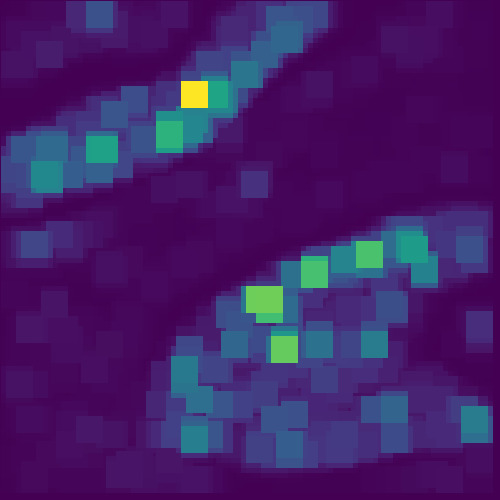}
            \caption{Epoch 120}
        \end{subfigure}
    \end{minipage}
    \caption{Visualization of the attention distribution when training with
             varying number of patches. All the distributions converge to
             approximately the same after $\sim$100 epochs.}
    \label{fig:crch_patches_ablation}
\end{figure*}

In Figure~\ref{fig:crch_patches_ablation}, we visualize, in a similar fashion
to \ref{sec:entropy_ablation}, the attention distributions learned when
sampling various numbers of patches per image for training. Although the
distributions are different in the beginning of training after approximately
100 epochs they converge to a very similar attention distribution.

\section{Qualitative results of the learned attention distribution}
\label{sec:qualitative}

In this section, we provide additional visualizations of the learned attention
distribution using both \ats{} and \mil{} on our two real world datasets,
namely the Histopathology images \S~\ref{sec:crch} and the Speed limits
\S~\ref{sec:speedlimits}.

\subsection{Histopathology images}\label{sec:crch}

In Figure~\ref{fig:crch_attention} we visualize the learned attention
distribution of \ats{} and we compare it to \mil{} and the ground truth
positions of epithelial cells in an subset of the test set.

\begin{figure*}
    \centering
    \begin{subfigure}[b]{0.24\columnwidth}
        \includegraphics[width=\textwidth]{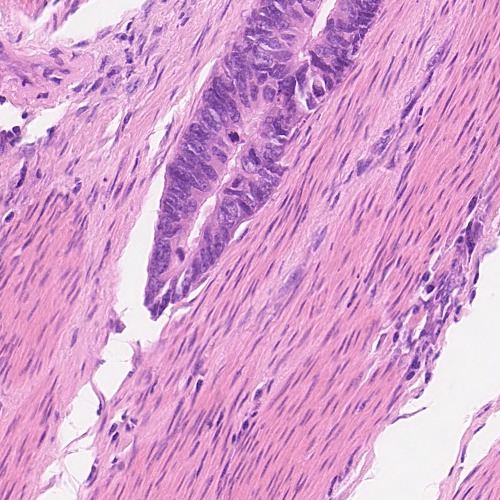}
        \caption{}
    \end{subfigure}
    \begin{subfigure}[b]{0.24\columnwidth}
        \includegraphics[width=\textwidth]{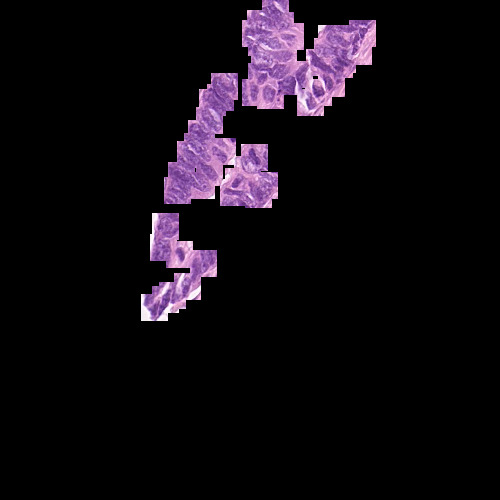}
        \caption{}
    \end{subfigure}
    \begin{subfigure}[b]{0.24\columnwidth}
        \includegraphics[width=\textwidth]{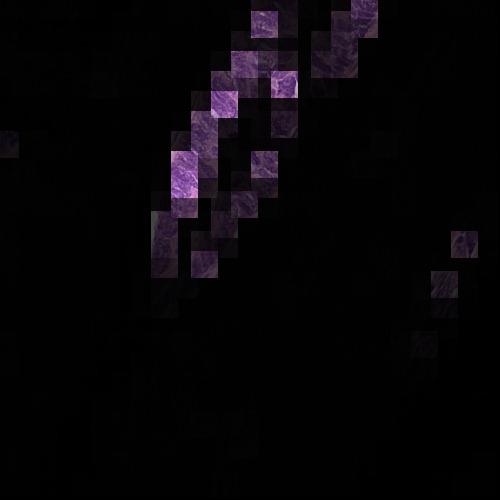}
        \caption{}
    \end{subfigure}
    \begin{subfigure}[b]{0.24\columnwidth}
        \includegraphics[width=\textwidth]{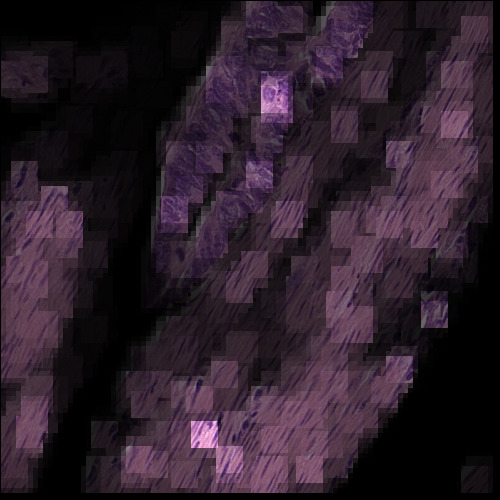}
        \caption{}
    \end{subfigure}
    \begin{subfigure}[b]{0.24\columnwidth}
        \includegraphics[width=\textwidth]{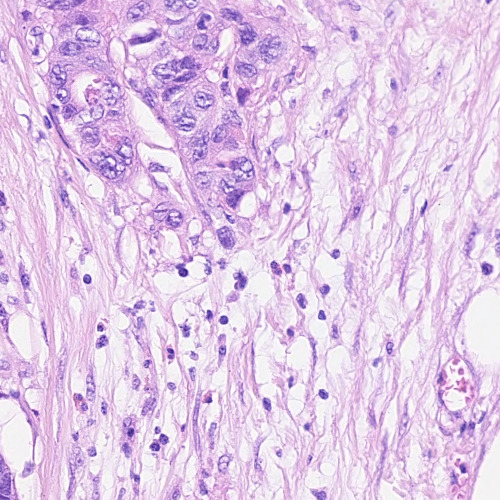}
        \caption{}
    \end{subfigure}
    \begin{subfigure}[b]{0.24\columnwidth}
        \includegraphics[width=\textwidth]{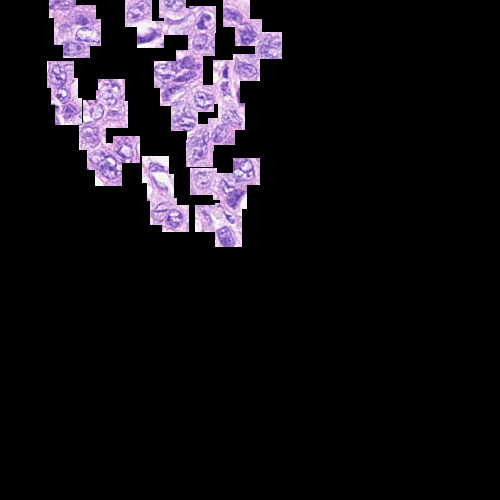}
        \caption{}
    \end{subfigure}
    \begin{subfigure}[b]{0.24\columnwidth}
        \includegraphics[width=\textwidth]{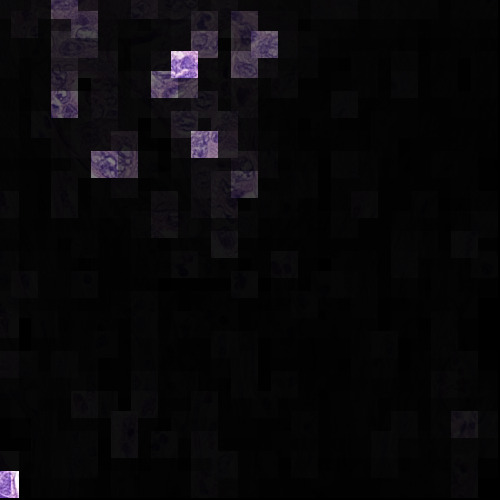}
        \caption{}
    \end{subfigure}
    \begin{subfigure}[b]{0.24\columnwidth}
        \includegraphics[width=\textwidth]{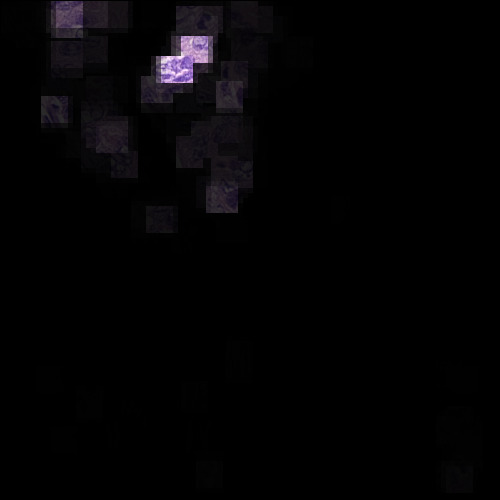}
        \caption{}
    \end{subfigure}
    \begin{subfigure}[b]{0.24\columnwidth}
        \includegraphics[width=\textwidth]{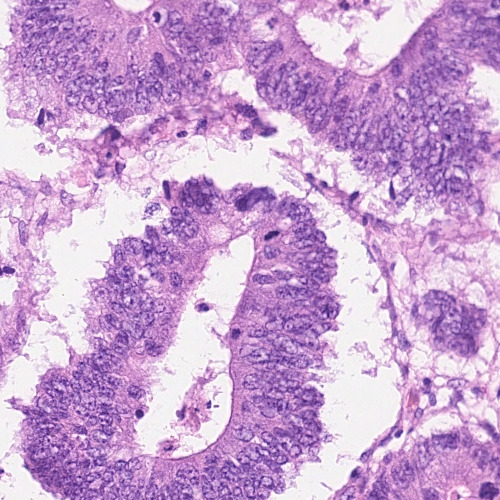}
        \caption{}
    \end{subfigure}
    \begin{subfigure}[b]{0.24\columnwidth}
        \includegraphics[width=\textwidth]{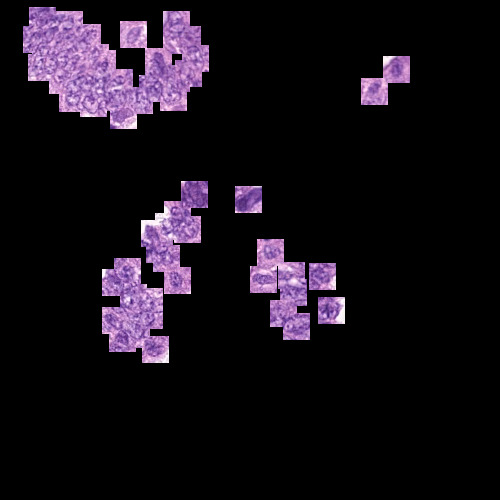}
        \caption{}
    \end{subfigure}
    \begin{subfigure}[b]{0.24\columnwidth}
        \includegraphics[width=\textwidth]{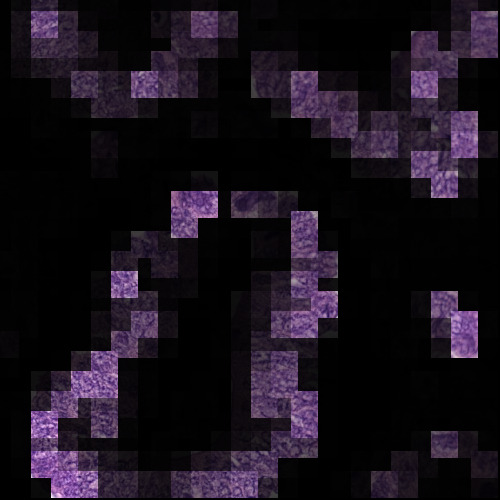}
        \caption{} \label{fig:crch_attention:mis1}
    \end{subfigure}
    \begin{subfigure}[b]{0.24\columnwidth}
        \includegraphics[width=\textwidth]{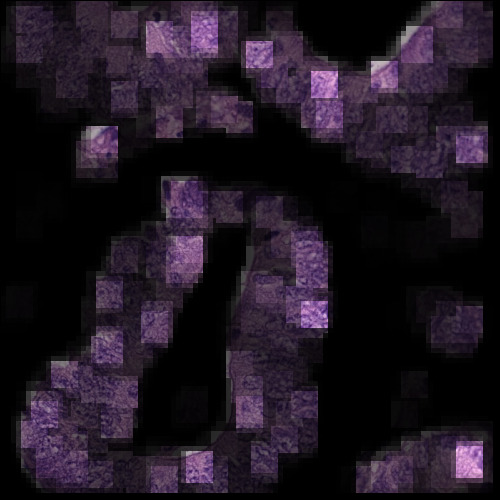}
        \caption{} \label{fig:crch_attention:mis2}
    \end{subfigure}
    \begin{subfigure}[b]{0.24\columnwidth}
        \includegraphics[width=\textwidth]{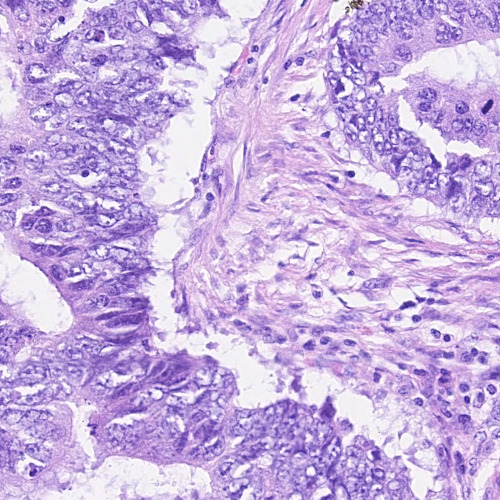}
        \caption{}
    \end{subfigure}
    \begin{subfigure}[b]{0.24\columnwidth}
        \includegraphics[width=\textwidth]{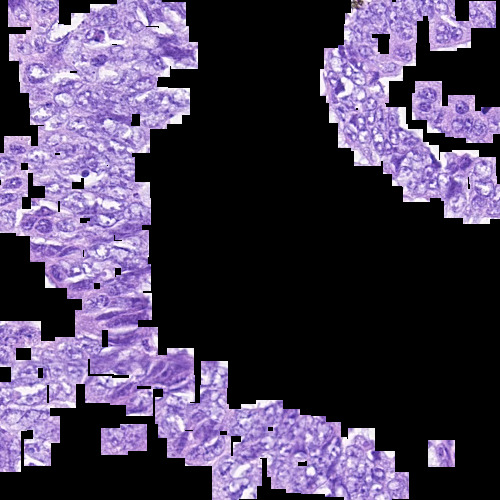}
        \caption{}
    \end{subfigure}
    \begin{subfigure}[b]{0.24\columnwidth}
        \includegraphics[width=\textwidth]{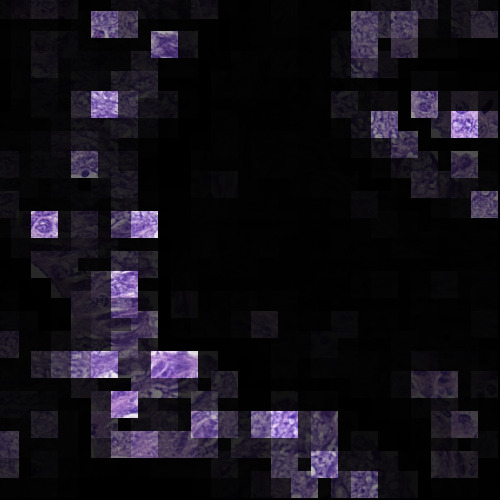}
        \caption{}
    \end{subfigure}
    \begin{subfigure}[b]{0.24\columnwidth}
        \includegraphics[width=\textwidth]{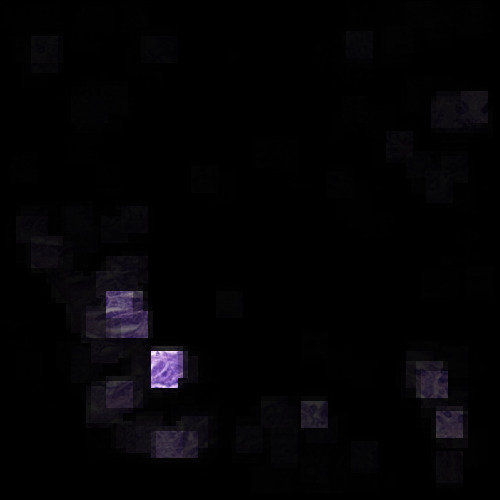}
        \caption{}
    \end{subfigure}
    \begin{subfigure}[b]{0.24\columnwidth}
        \includegraphics[width=\textwidth]{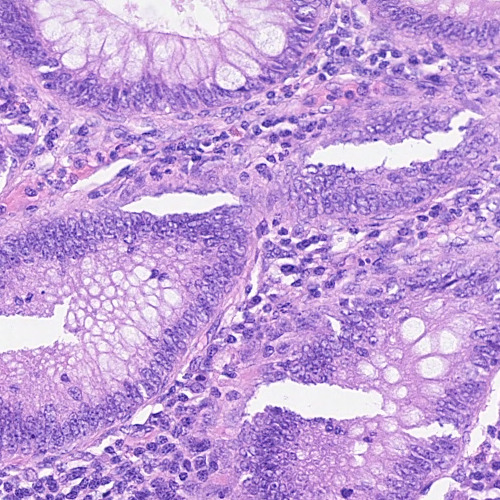}
        \caption{}
    \end{subfigure}
    \begin{subfigure}[b]{0.24\columnwidth}
        \includegraphics[width=\textwidth]{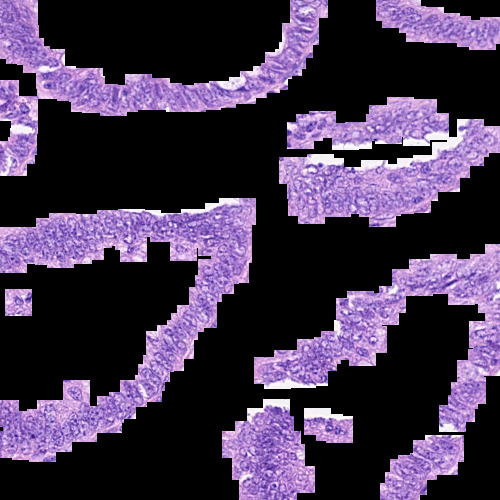}
        \caption{}
    \end{subfigure}
    \begin{subfigure}[b]{0.24\columnwidth}
        \includegraphics[width=\textwidth]{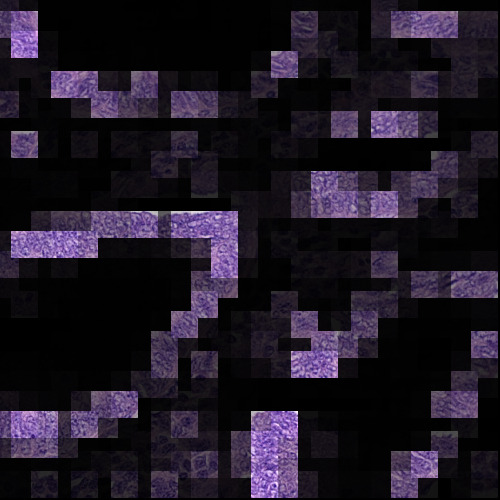}
        \caption{}
    \end{subfigure}
    \begin{subfigure}[b]{0.24\columnwidth}
        \includegraphics[width=\textwidth]{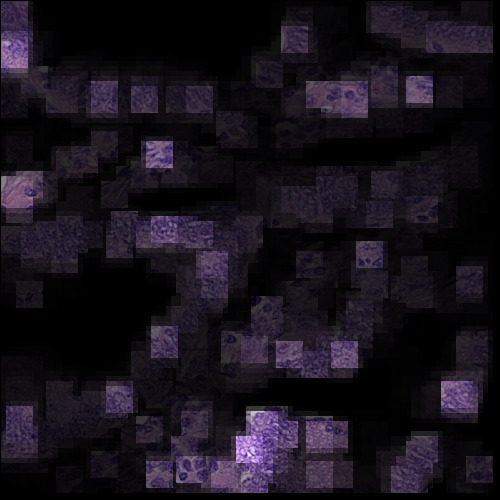}
        \caption{}
    \end{subfigure}
    \begin{subfigure}[b]{0.24\columnwidth}
        \includegraphics[width=\textwidth]{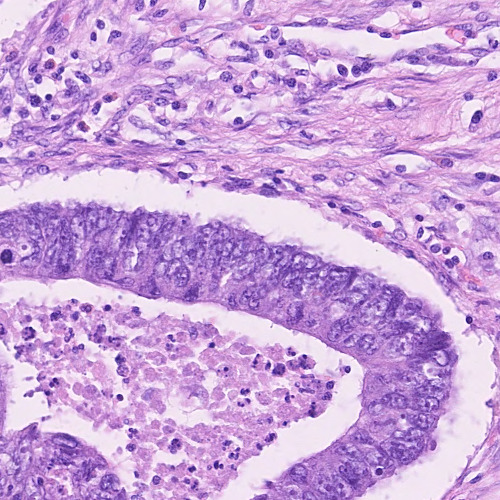}
        \caption{}
    \end{subfigure}
    \begin{subfigure}[b]{0.24\columnwidth}
        \includegraphics[width=\textwidth]{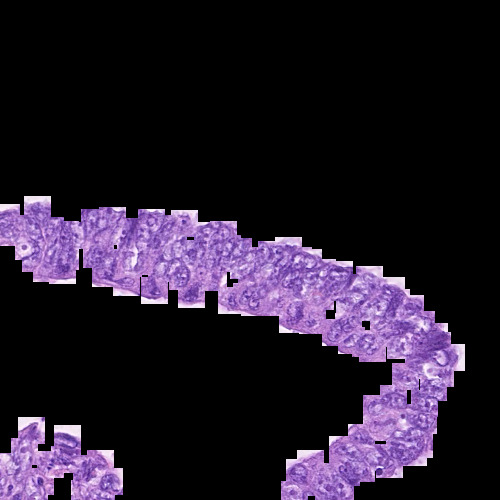}
        \caption{}
    \end{subfigure}
    \begin{subfigure}[b]{0.24\columnwidth}
        \includegraphics[width=\textwidth]{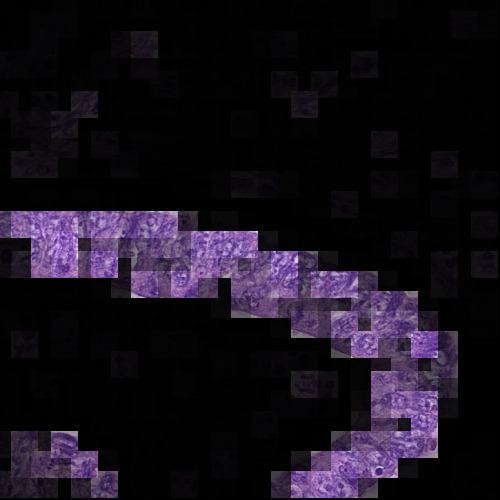}
        \caption{}
    \end{subfigure}
    \begin{subfigure}[b]{0.24\columnwidth}
        \includegraphics[width=\textwidth]{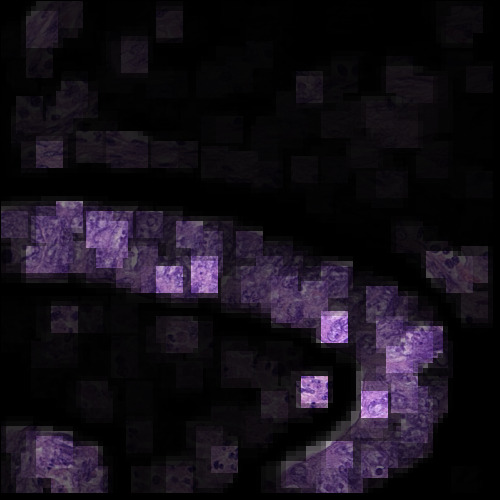}
        \caption{}
    \end{subfigure}
    \begin{subfigure}[b]{0.24\columnwidth}
        \includegraphics[width=\textwidth]{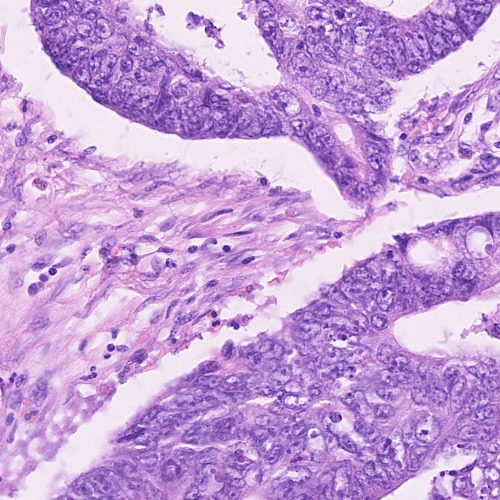}
        \caption{}
    \end{subfigure}
    \begin{subfigure}[b]{0.24\columnwidth}
        \includegraphics[width=\textwidth]{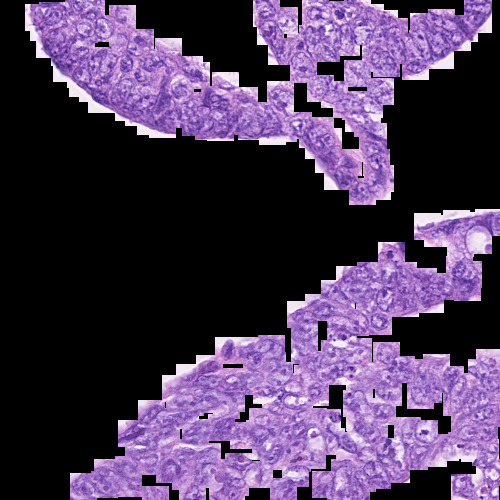}
        \caption{}
    \end{subfigure}
    \begin{subfigure}[b]{0.24\columnwidth}
        \includegraphics[width=\textwidth]{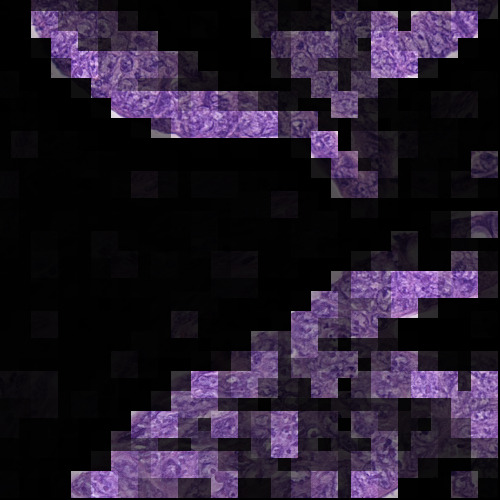}
        \caption{}
    \end{subfigure}
    \begin{subfigure}[b]{0.24\columnwidth}
        \includegraphics[width=\textwidth]{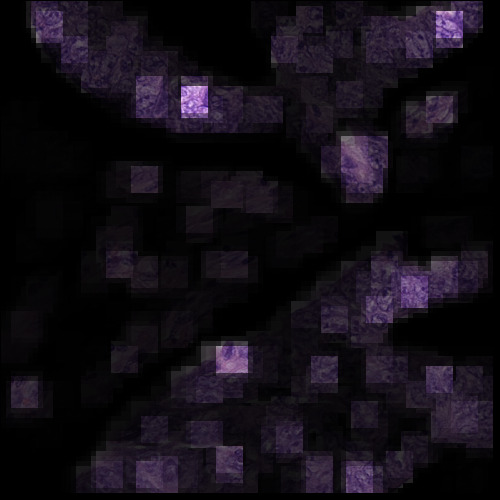}
        \caption{}
    \end{subfigure}
    \begin{subfigure}[b]{0.24\columnwidth}
        \includegraphics[width=\textwidth]{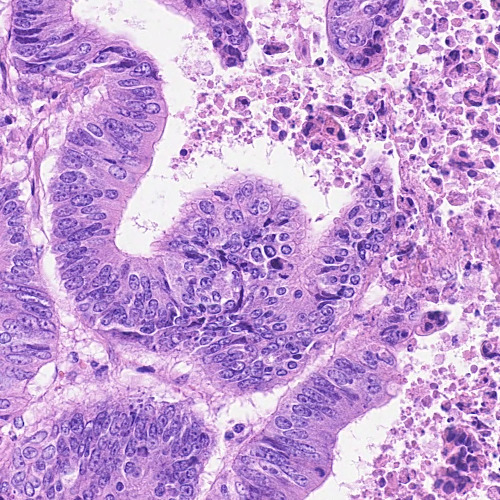}
        \caption{}
    \end{subfigure}
    \begin{subfigure}[b]{0.24\columnwidth}
        \includegraphics[width=\textwidth]{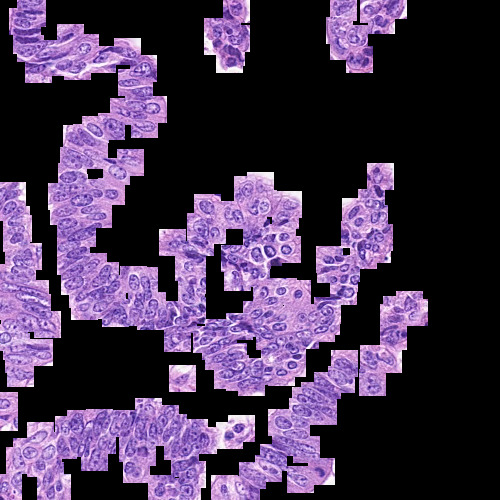}
        \caption{}
    \end{subfigure}
    \begin{subfigure}[b]{0.24\columnwidth}
        \includegraphics[width=\textwidth]{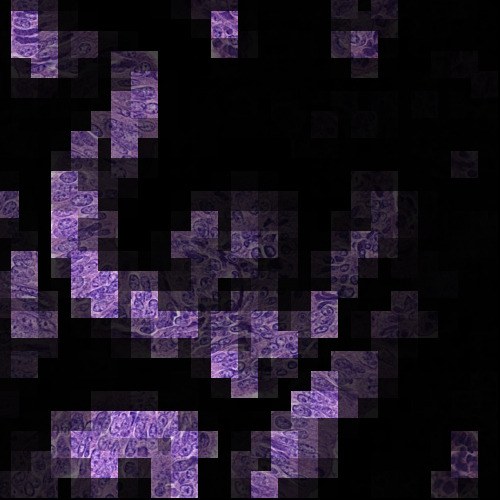}
        \caption{}
    \end{subfigure}
    \begin{subfigure}[b]{0.24\columnwidth}
        \includegraphics[width=\textwidth]{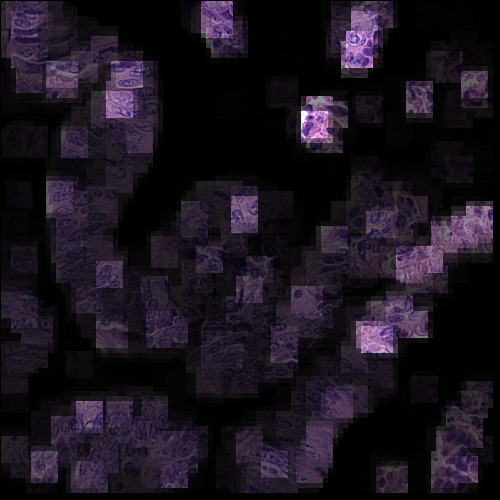}
        \caption{}
    \end{subfigure}
    \begin{subfigure}[b]{0.24\columnwidth}
        \includegraphics[width=\textwidth]{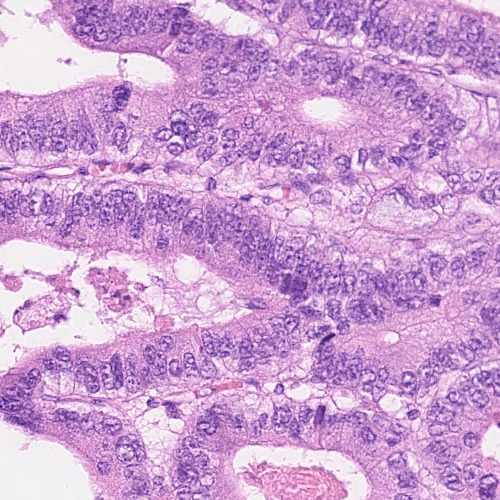}
        \caption{}
    \end{subfigure}
    \begin{subfigure}[b]{0.24\columnwidth}
        \includegraphics[width=\textwidth]{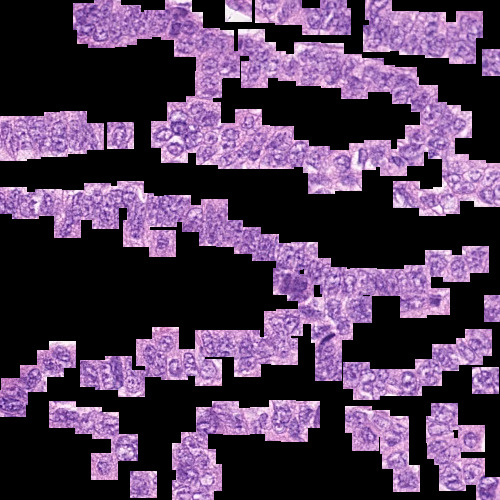}
        \caption{}
    \end{subfigure}
    \begin{subfigure}[b]{0.24\columnwidth}
        \includegraphics[width=\textwidth]{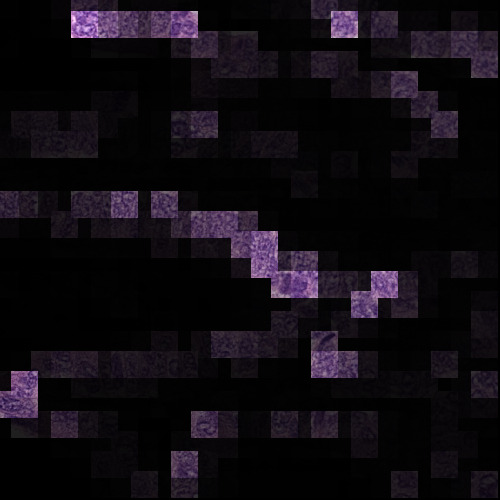}
        \caption{}
    \end{subfigure}
    \begin{subfigure}[b]{0.24\columnwidth}
        \includegraphics[width=\textwidth]{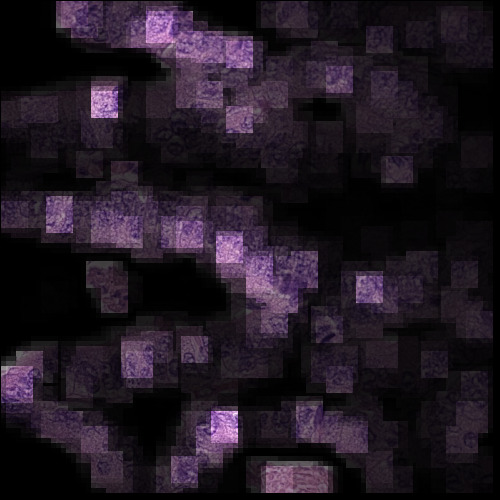}
        \caption{}
    \end{subfigure}
    \begin{subfigure}[b]{0.24\columnwidth}
        \includegraphics[width=\textwidth]{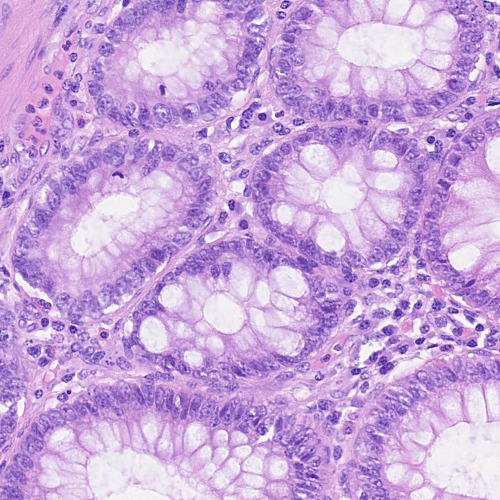}
        \caption{}
    \end{subfigure}
    \begin{subfigure}[b]{0.24\columnwidth}
        \includegraphics[width=\textwidth]{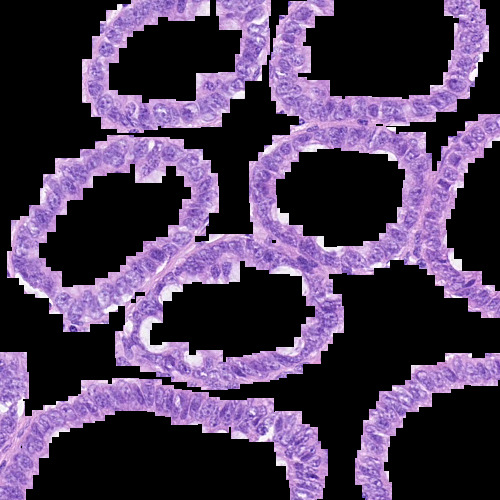}
        \caption{}
    \end{subfigure}
    \begin{subfigure}[b]{0.24\columnwidth}
        \includegraphics[width=\textwidth]{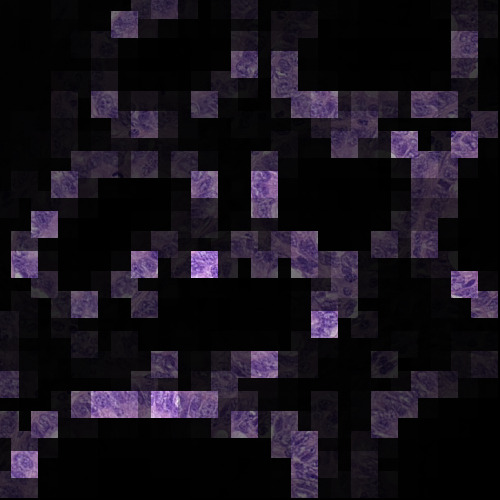}
        \caption{}
    \end{subfigure}
    \begin{subfigure}[b]{0.24\columnwidth}
        \includegraphics[width=\textwidth]{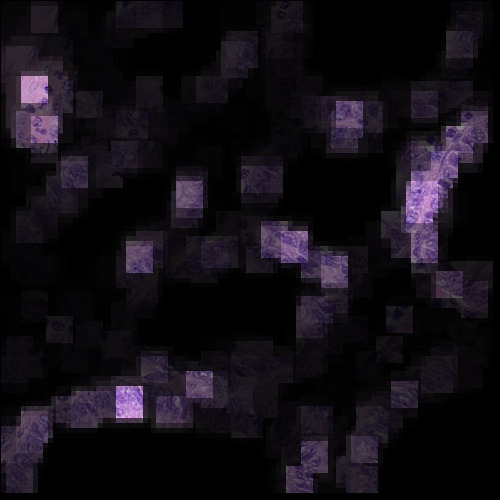}
        \caption{}
    \end{subfigure}
    \begin{subfigure}[b]{0.24\columnwidth}
        \includegraphics[width=\textwidth]{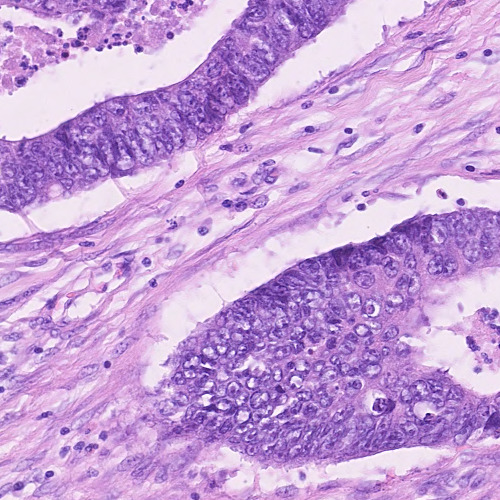}
        \caption{}
    \end{subfigure}
    \begin{subfigure}[b]{0.24\columnwidth}
        \includegraphics[width=\textwidth]{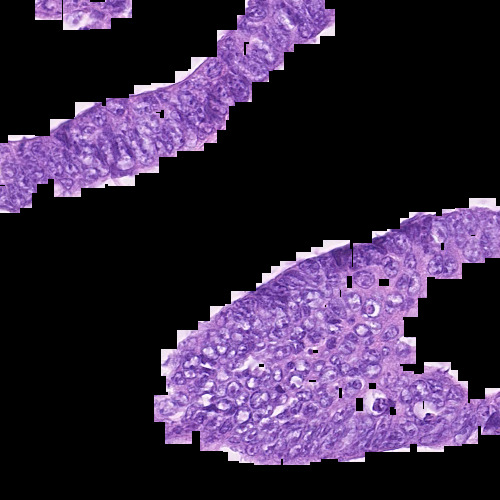}
        \caption{}
    \end{subfigure}
    \begin{subfigure}[b]{0.24\columnwidth}
        \includegraphics[width=\textwidth]{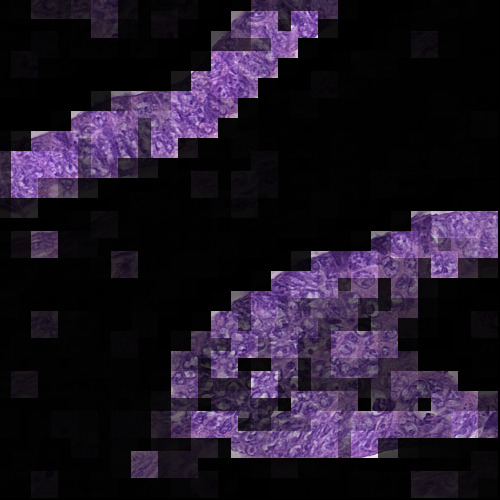}
        \caption{}
    \end{subfigure}
    \begin{subfigure}[b]{0.24\columnwidth}
        \includegraphics[width=\textwidth]{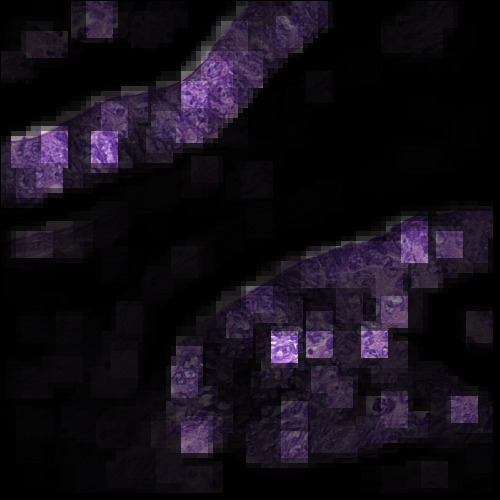}
        \caption{}
    \end{subfigure}
    \begin{subfigure}[b]{0.24\columnwidth}
        \includegraphics[width=\textwidth]{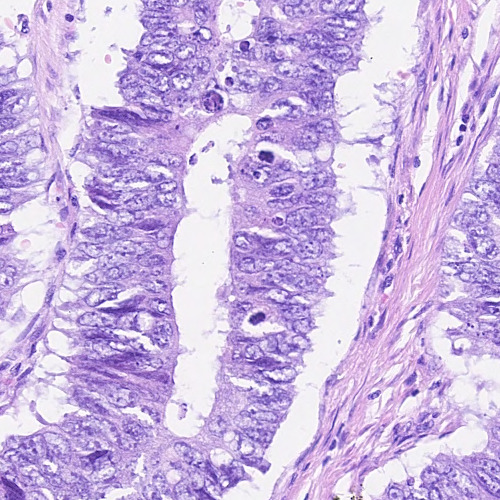}
        \caption{}
    \end{subfigure}
    \begin{subfigure}[b]{0.24\columnwidth}
        \includegraphics[width=\textwidth]{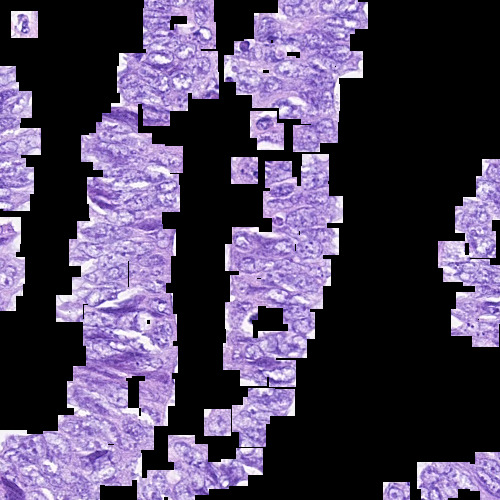}
        \caption{}
    \end{subfigure}
    \begin{subfigure}[b]{0.24\columnwidth}
        \includegraphics[width=\textwidth]{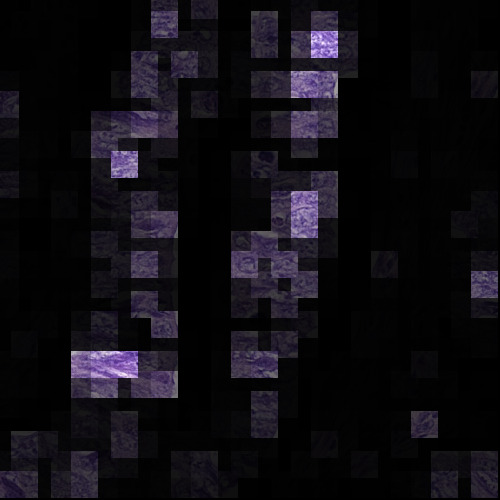}
        \caption{}
    \end{subfigure}
    \begin{subfigure}[b]{0.24\columnwidth}
        \includegraphics[width=\textwidth]{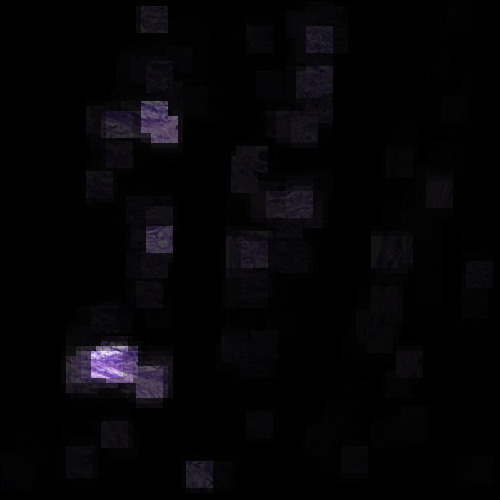}
        \caption{}
    \end{subfigure}
    \caption{We visualize in groups of 4, the H\&E stained image, the ground
             truth positions of epithelial cells, the attention distribution of
             \mil{} and the attention distribution of \ats{}. We observe that
             indeed our method learns to identify regions of interest without
             per patch annotations in a similar fashion to \mil{}.}
    \label{fig:crch_attention}
\end{figure*}

We observe that the learned attention distribution is very similar to the one
learned by \mil{} even though our model processes a fraction of the image at
any iteration. In addition, it is interesting to note that the two methods
produce distributions that agree even on mistakenly tagged patches, one such
case is depicted in figures \subref{fig:crch_attention:mis1} and
\subref{fig:crch_attention:mis2} where both methods the top right part of the
image to contain useful patches.

\subsection{Speed limits}\label{sec:speedlimits}

\begin{figure*}
    \centering

    %

    \begin{subfigure}[b]{0.25\textwidth}
        \includegraphics[width=\textwidth]{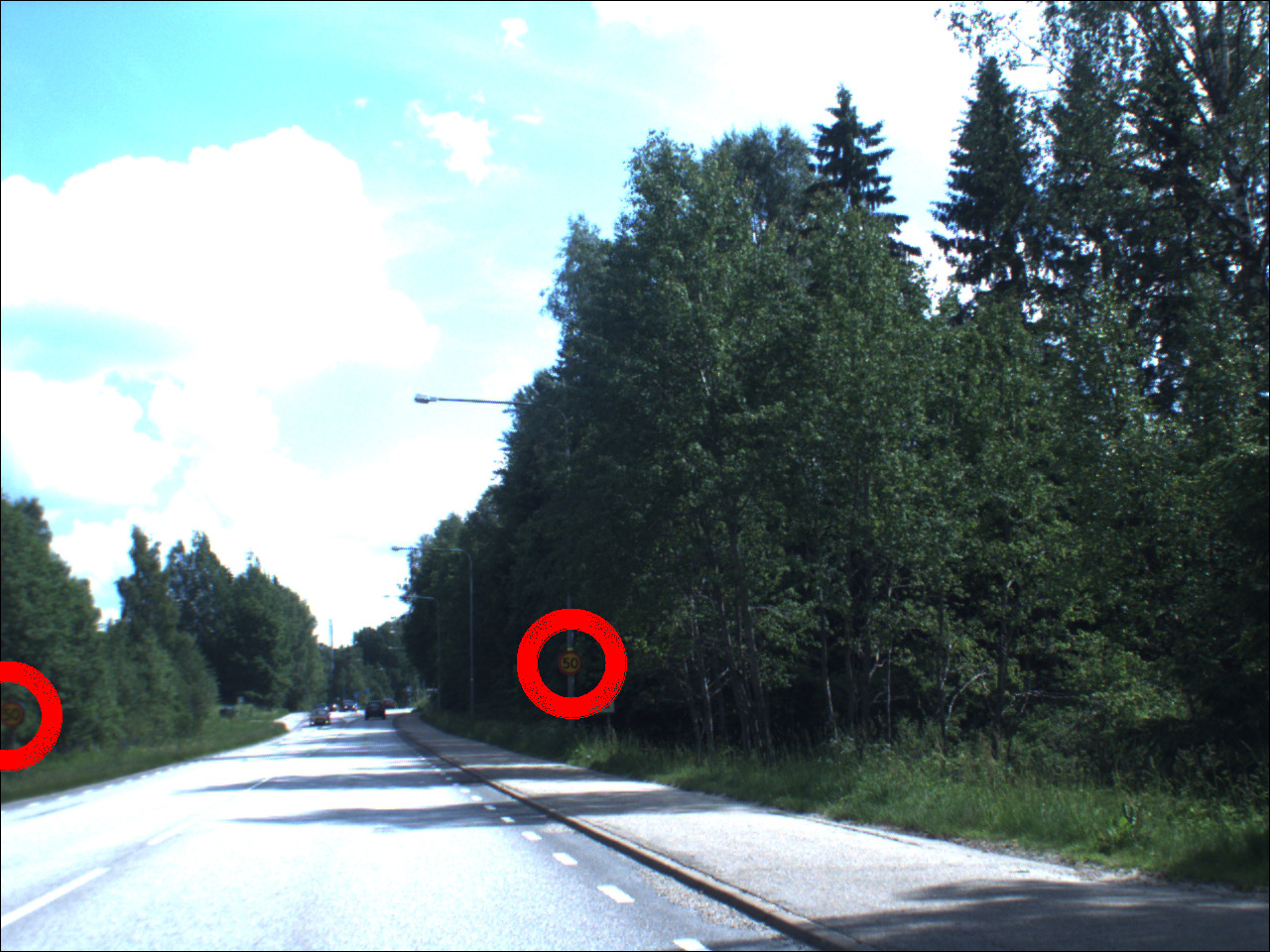}
        \caption{}
        \vspace{1.5em}
    \end{subfigure}
    \begin{subfigure}[b]{0.25\textwidth}
        \includegraphics[width=\textwidth]{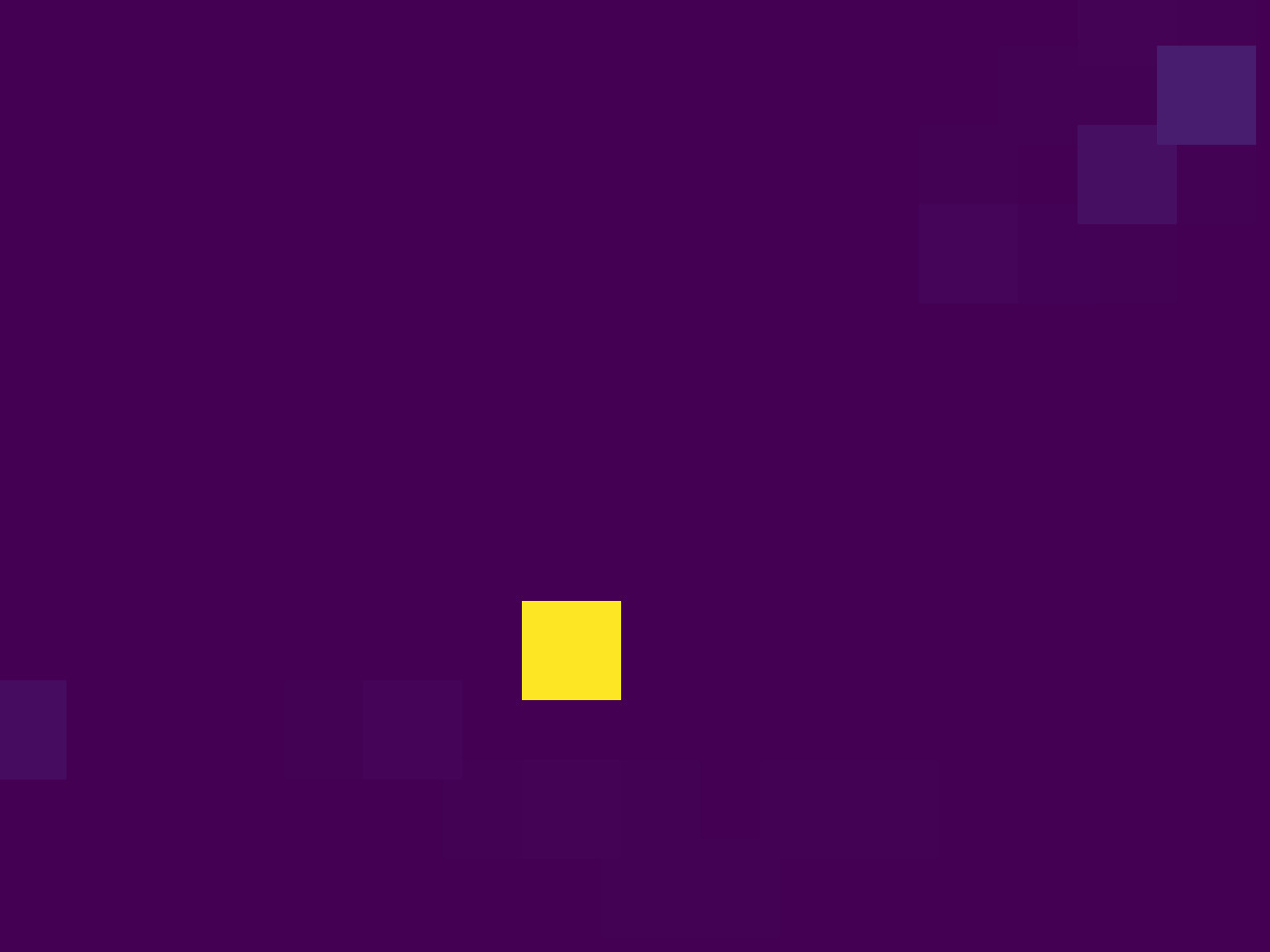}
        \caption{}
        \vspace{1.5em}
    \end{subfigure}
    \begin{subfigure}[b]{0.25\textwidth}
        \includegraphics[width=\textwidth]{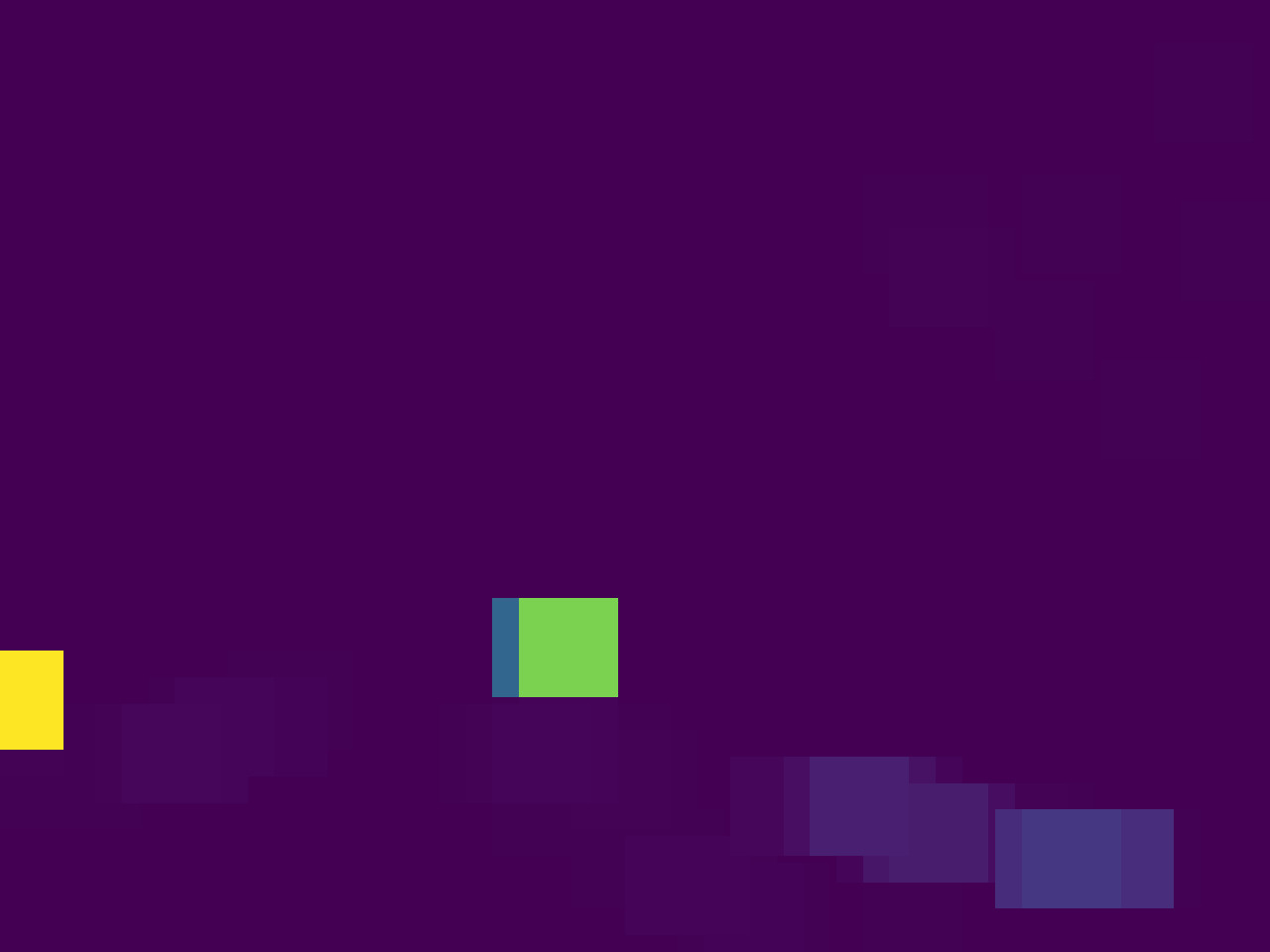}
        \caption{}
        \vspace{1.5em}
    \end{subfigure}
    \begin{minipage}[b]{0.11\textwidth}
        \begin{subfigure}[b]{\textwidth}
            \centering
            \includegraphics[width=\textwidth]{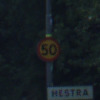}
            \caption{}
        \end{subfigure}
        \begin{subfigure}[b]{\textwidth}
            \centering
            \includegraphics[width=\textwidth]{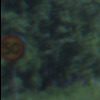}
            \caption{}
        \end{subfigure}
    \end{minipage}

    \begin{subfigure}[b]{0.25\textwidth}
        \includegraphics[width=\textwidth]{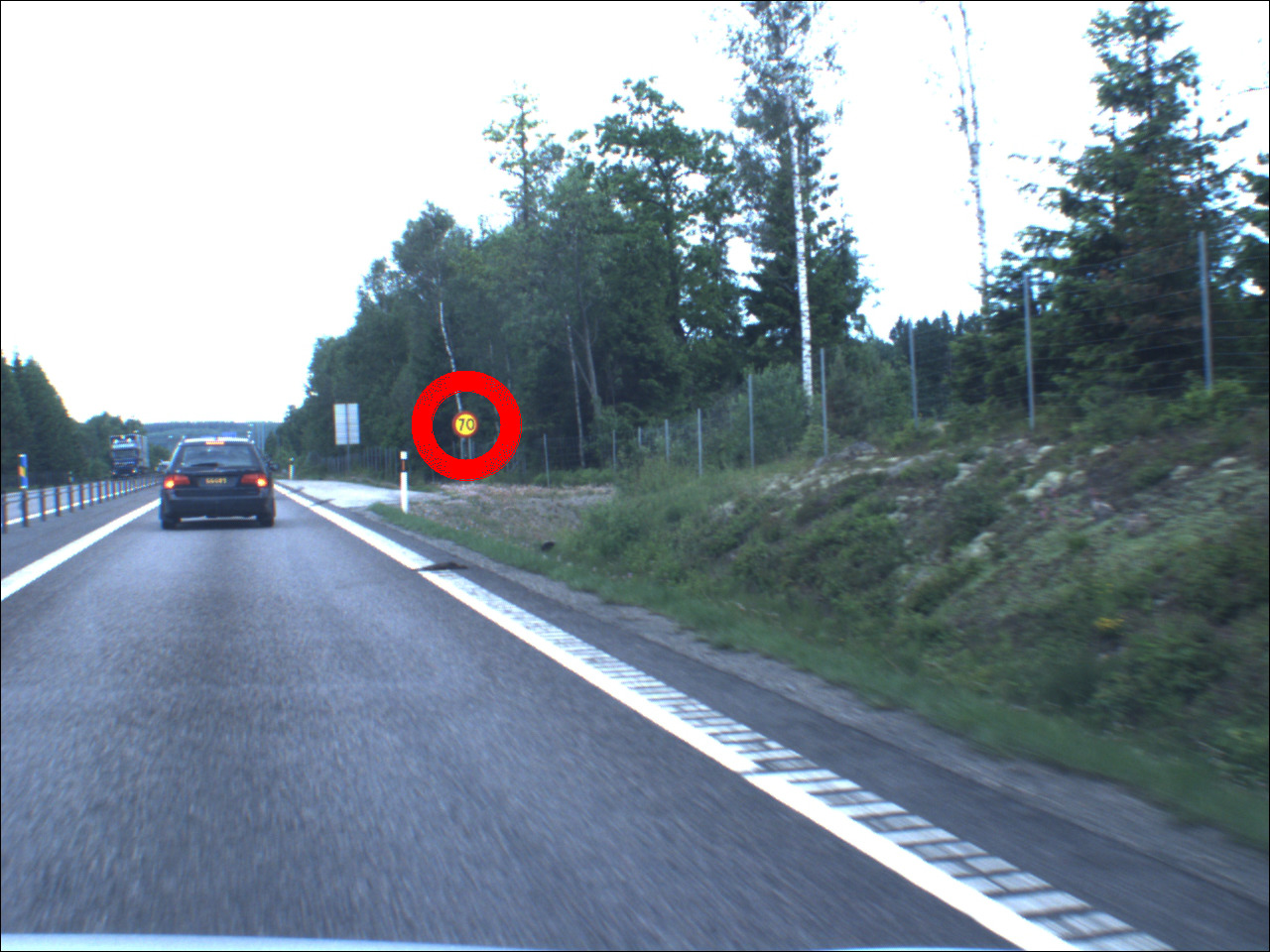}
        \caption{} \label{fig:speed_limits1}
    \end{subfigure}
    \begin{subfigure}[b]{0.25\textwidth}
        \includegraphics[width=\textwidth]{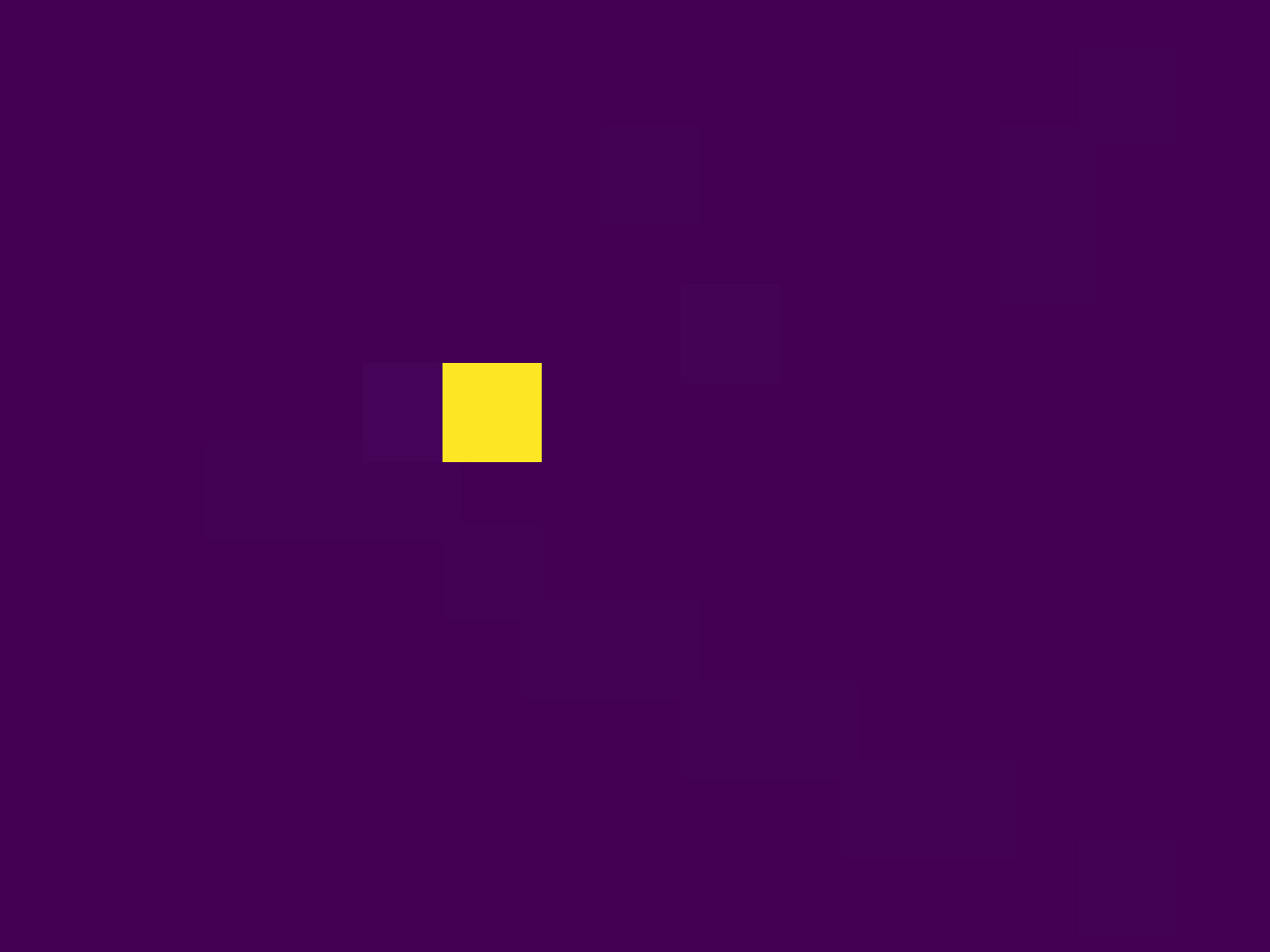}
        \caption{}
    \end{subfigure}
    \begin{subfigure}[b]{0.25\textwidth}
        \includegraphics[width=\textwidth]{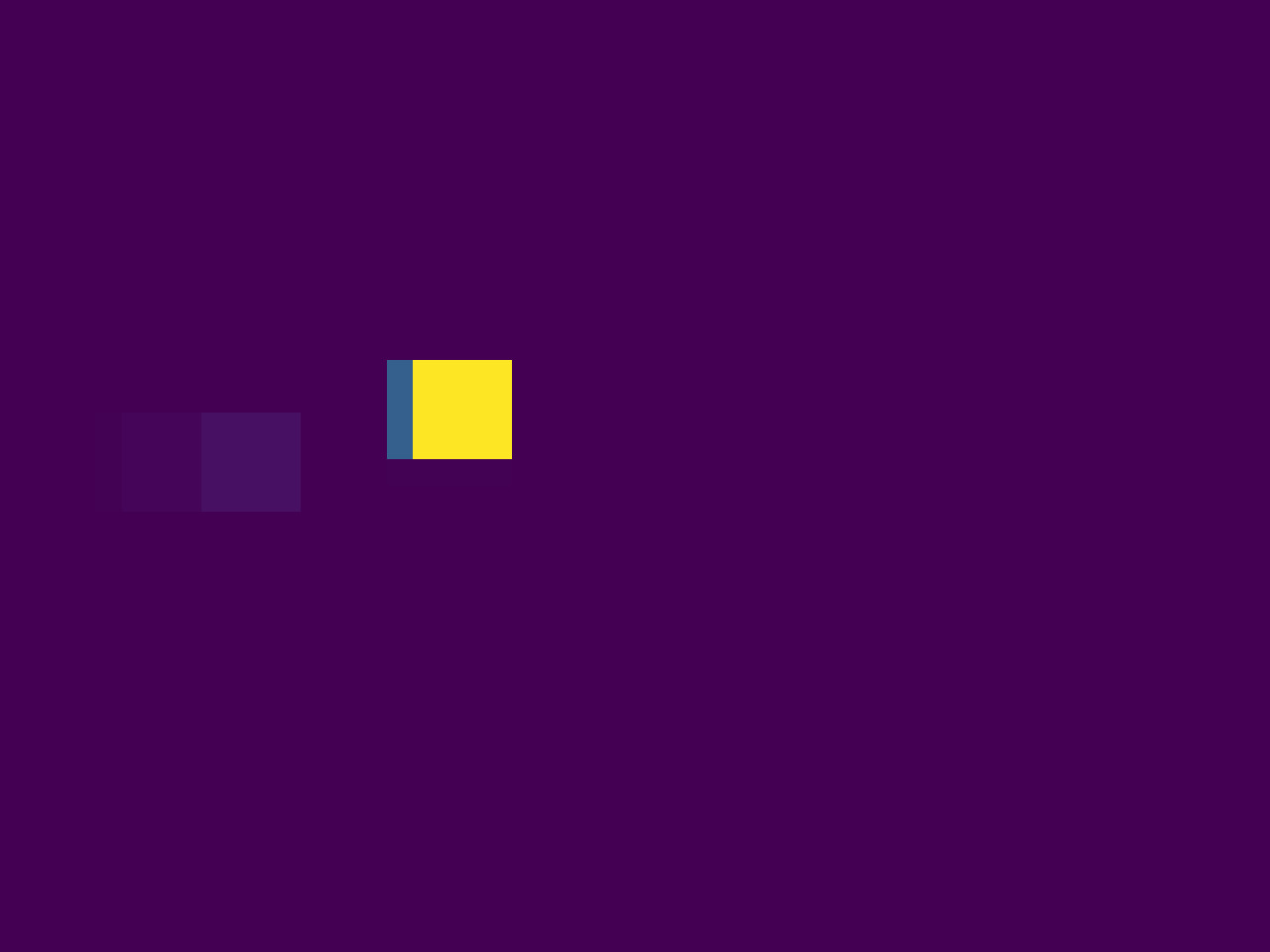}
        \caption{}
    \end{subfigure}
    \begin{minipage}[b]{0.11\textwidth}
        \begin{subfigure}[b]{\textwidth}
            \centering
            \includegraphics[width=\textwidth]{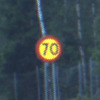}
            \caption{}
        \end{subfigure}
        \vspace{1em}
    \end{minipage}

    \begin{subfigure}[b]{0.25\textwidth}
        \includegraphics[width=\textwidth]{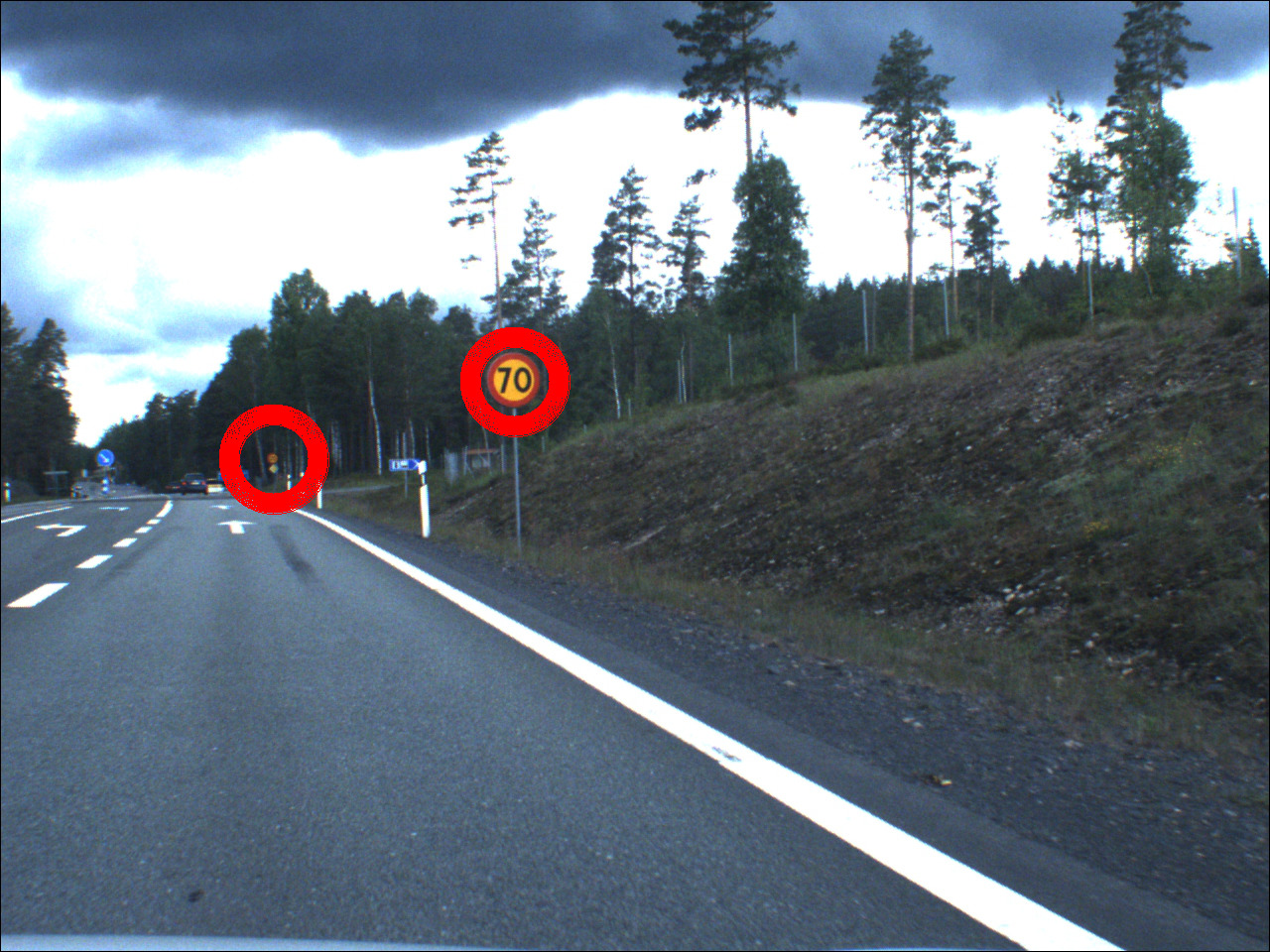}
        \caption{}
        \vspace{1.5em}
    \end{subfigure}
    \begin{subfigure}[b]{0.25\textwidth}
        \includegraphics[width=\textwidth]{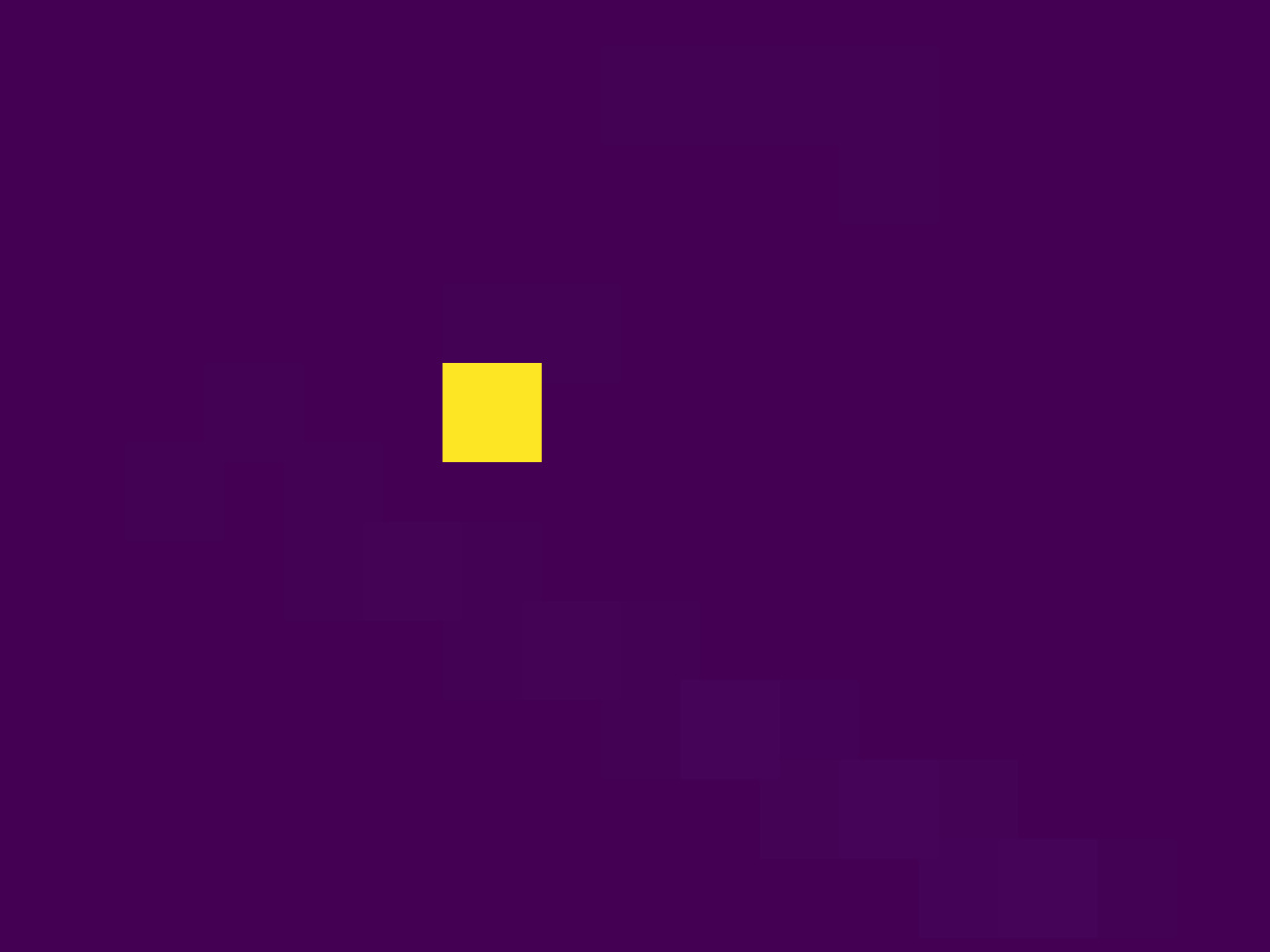}
        \caption{}
        \vspace{1.5em}
    \end{subfigure}
    \begin{subfigure}[b]{0.25\textwidth}
        \includegraphics[width=\textwidth]{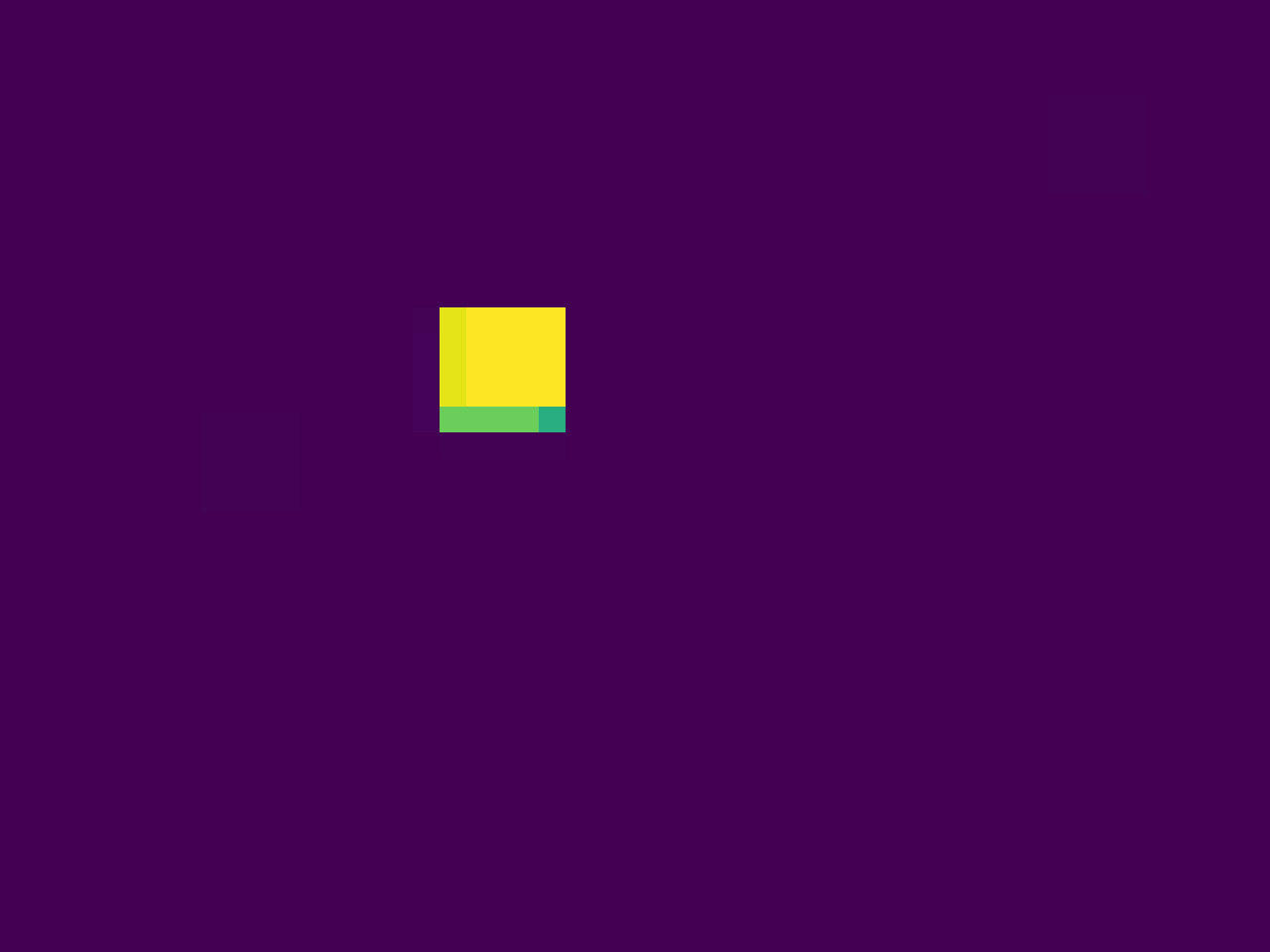}
        \caption{}
        \vspace{1.5em}
    \end{subfigure}
    \begin{minipage}[b]{0.11\textwidth}
        \begin{subfigure}[b]{\textwidth}
            \centering
            \includegraphics[width=\textwidth]{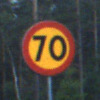}
            \caption{}
        \end{subfigure}
        \begin{subfigure}[b]{\textwidth}
            \centering
            \includegraphics[width=\textwidth]{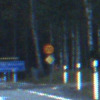}
            \caption{}
        \end{subfigure}
    \end{minipage}

    \begin{subfigure}[b]{0.25\textwidth}
        \includegraphics[width=\textwidth]{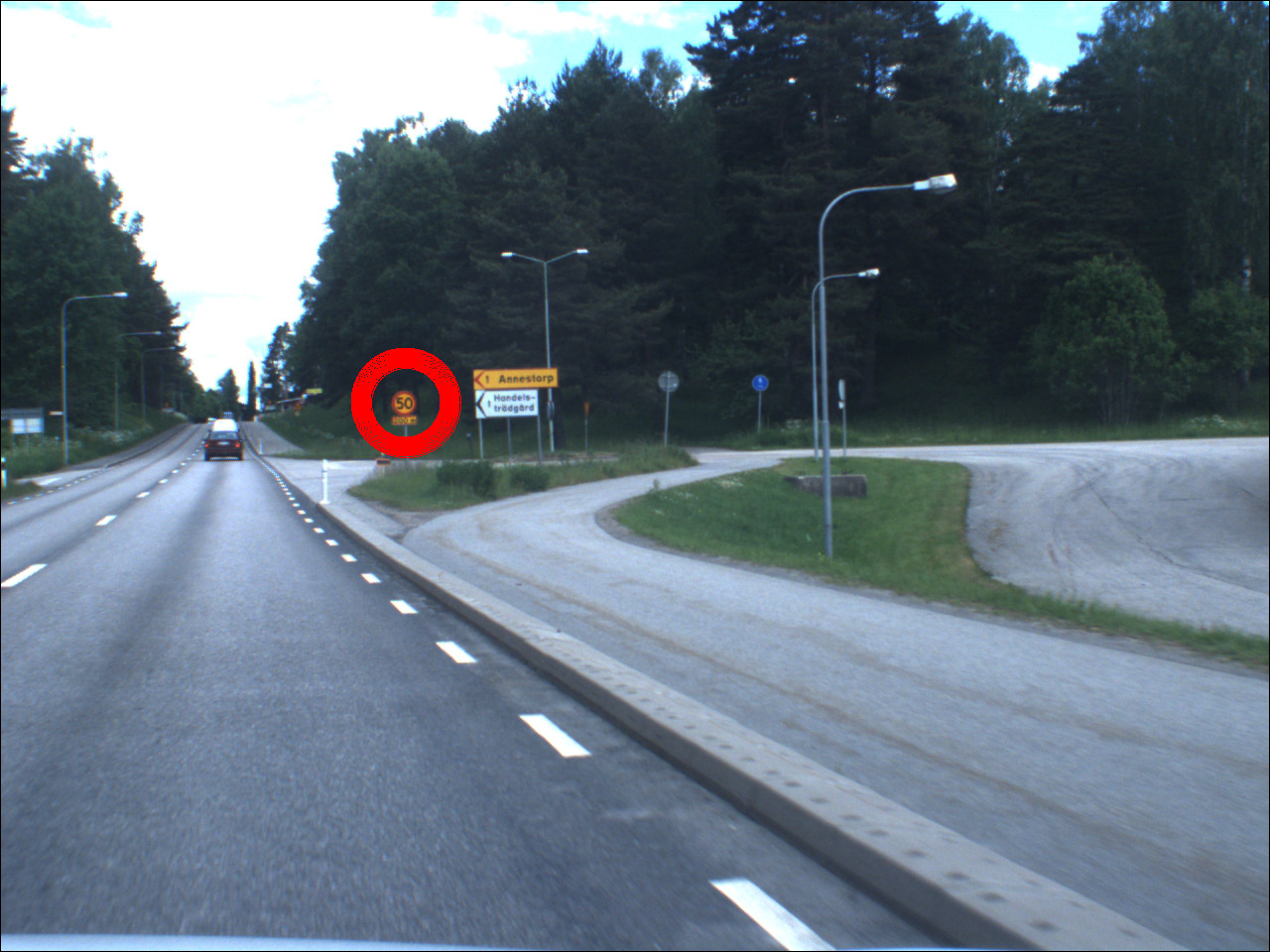}
        \caption{} \label{fig:speed_limits2}
    \end{subfigure}
    \begin{subfigure}[b]{0.25\textwidth}
        \includegraphics[width=\textwidth]{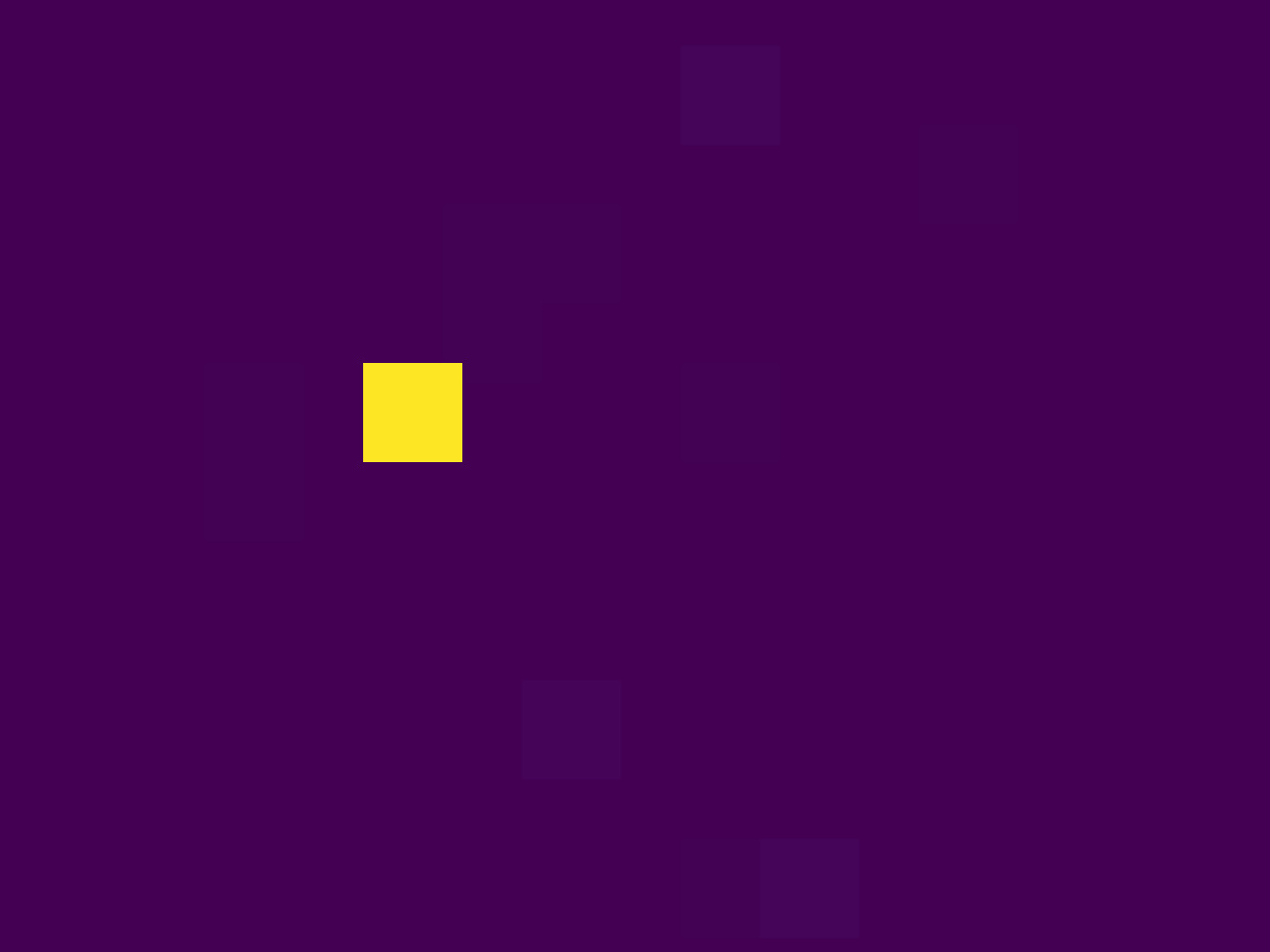}
        \caption{}
    \end{subfigure}
    \begin{subfigure}[b]{0.25\textwidth}
        \includegraphics[width=\textwidth]{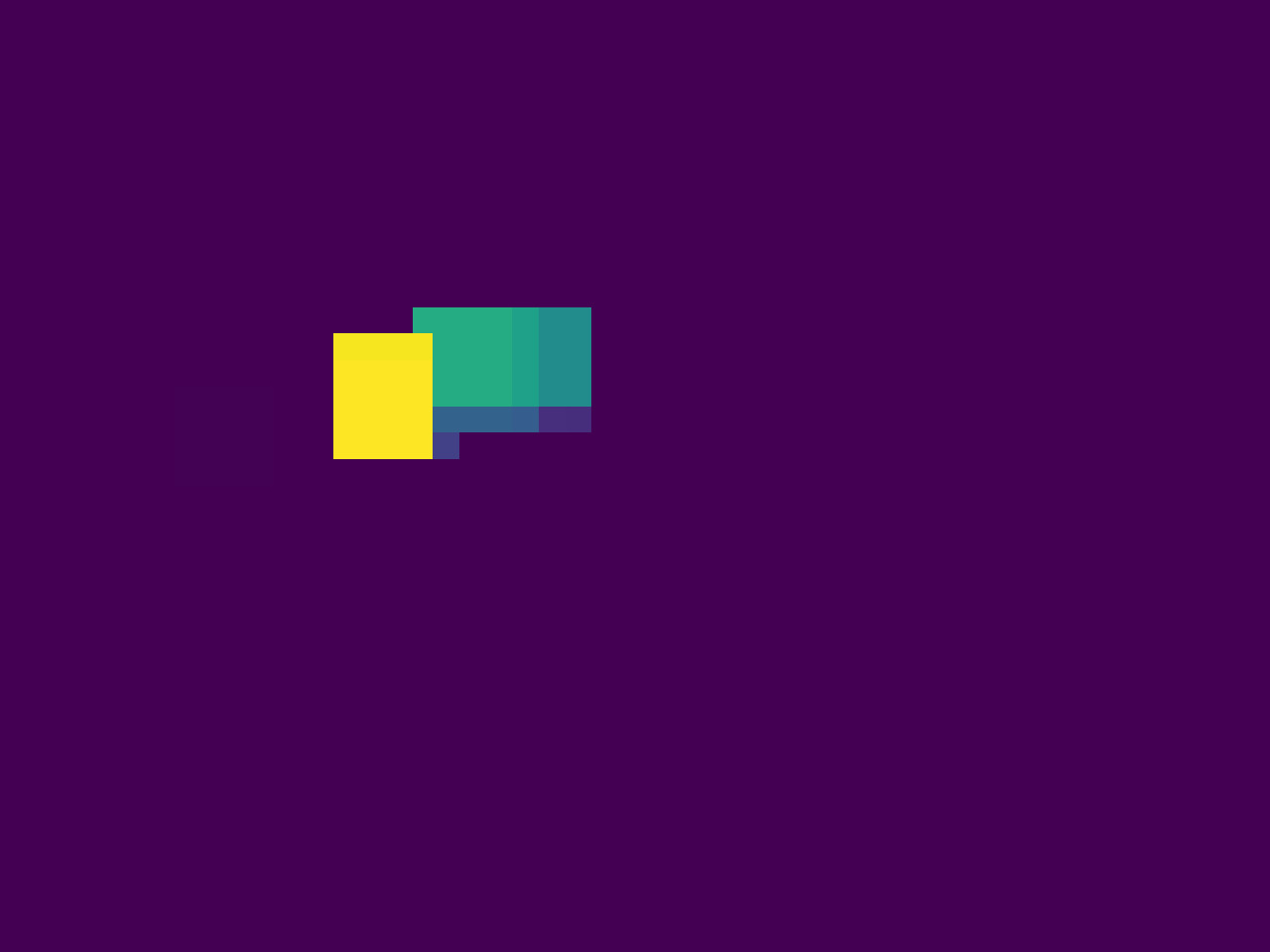}
        \caption{}
    \end{subfigure}
    \begin{minipage}[b]{0.11\textwidth}
        \begin{subfigure}[b]{\textwidth}
            \centering
            \includegraphics[width=\textwidth]{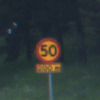}
            \caption{}
        \end{subfigure}
        \vspace{1em}
    \end{minipage}

    \begin{subfigure}[b]{0.25\textwidth}
        \includegraphics[width=\textwidth]{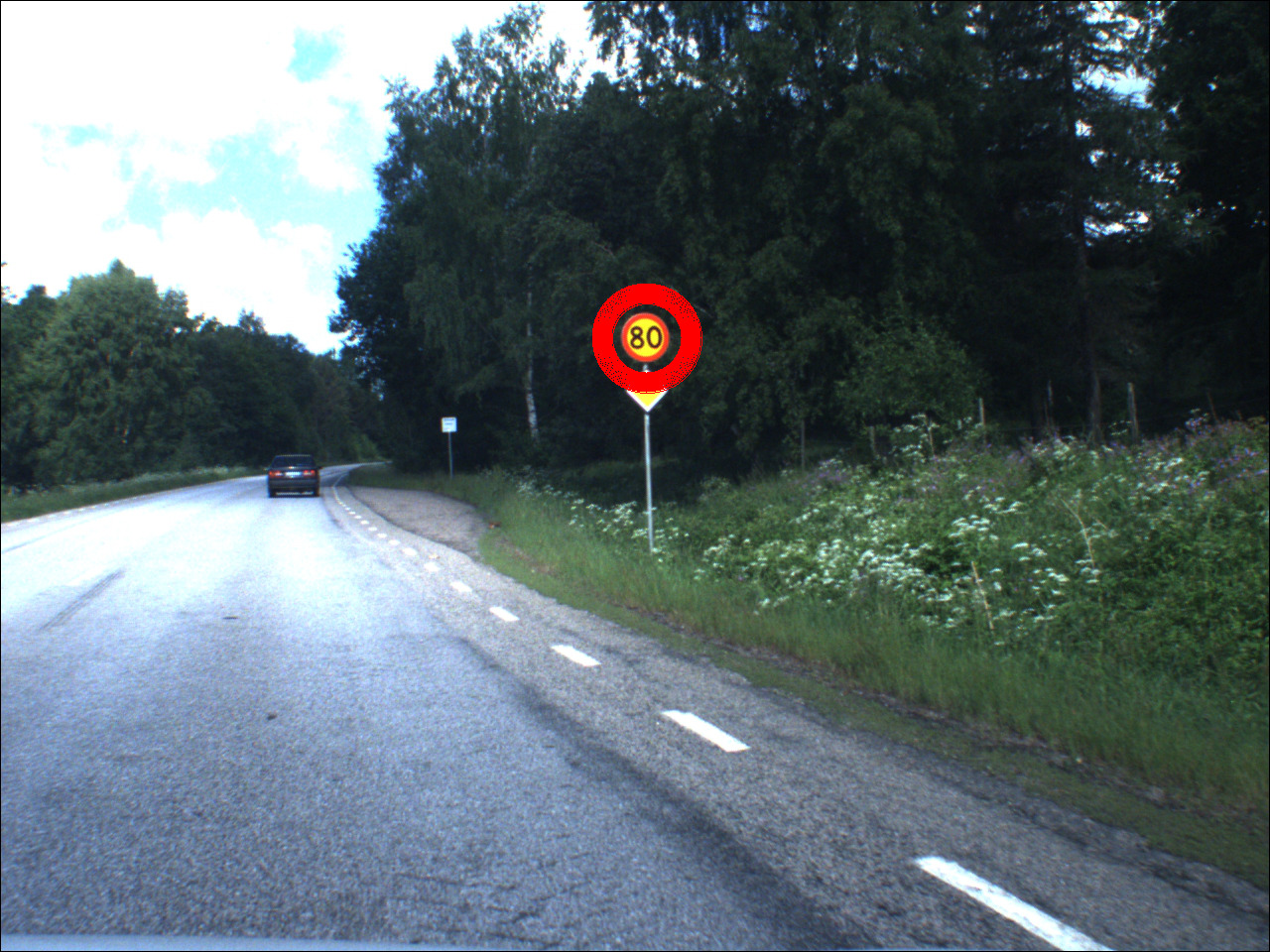}
        \caption{}
    \end{subfigure}
    \begin{subfigure}[b]{0.25\textwidth}
        \includegraphics[width=\textwidth]{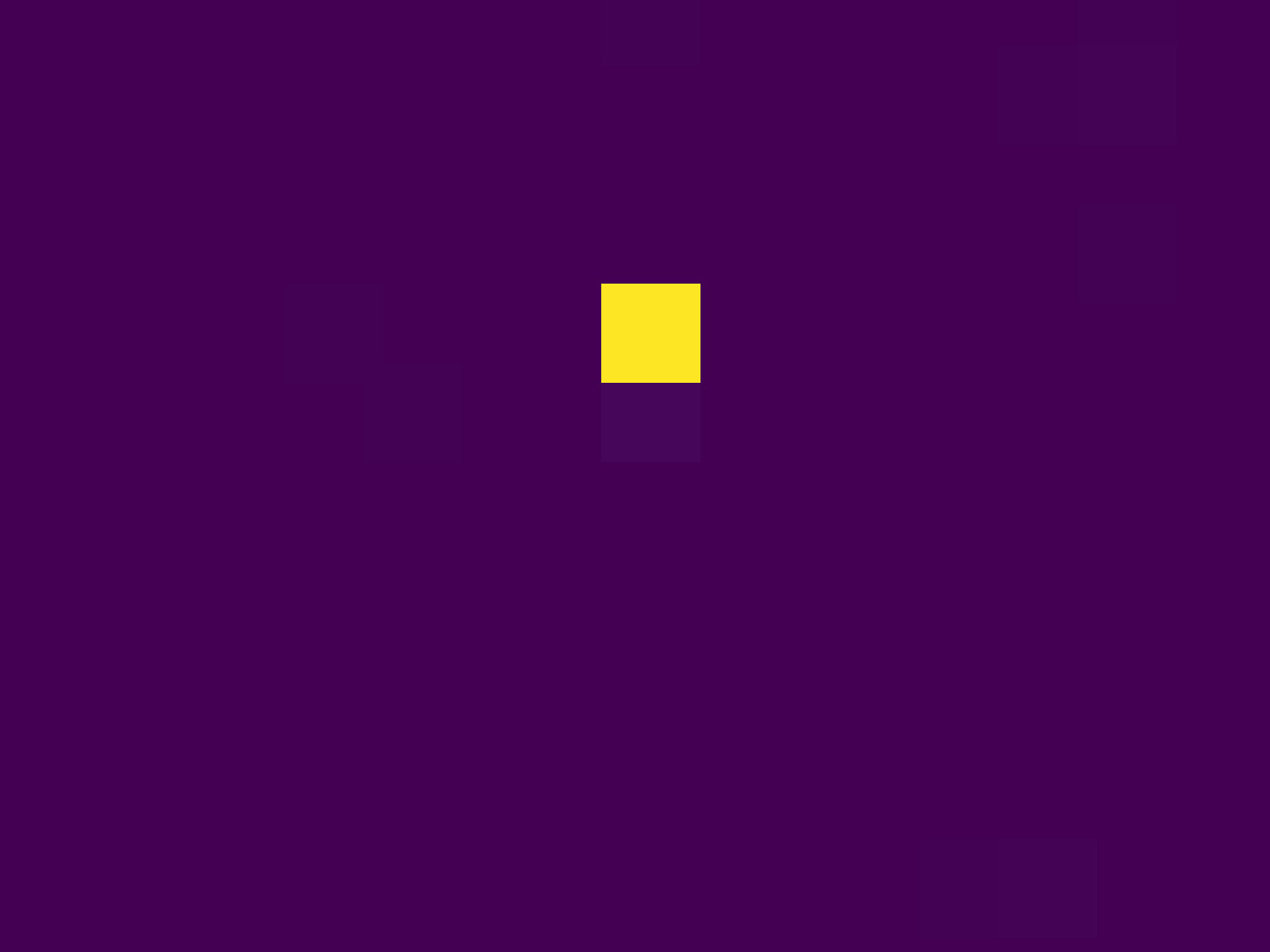}
        \caption{}
    \end{subfigure}
    \begin{subfigure}[b]{0.25\textwidth}
        \includegraphics[width=\textwidth]{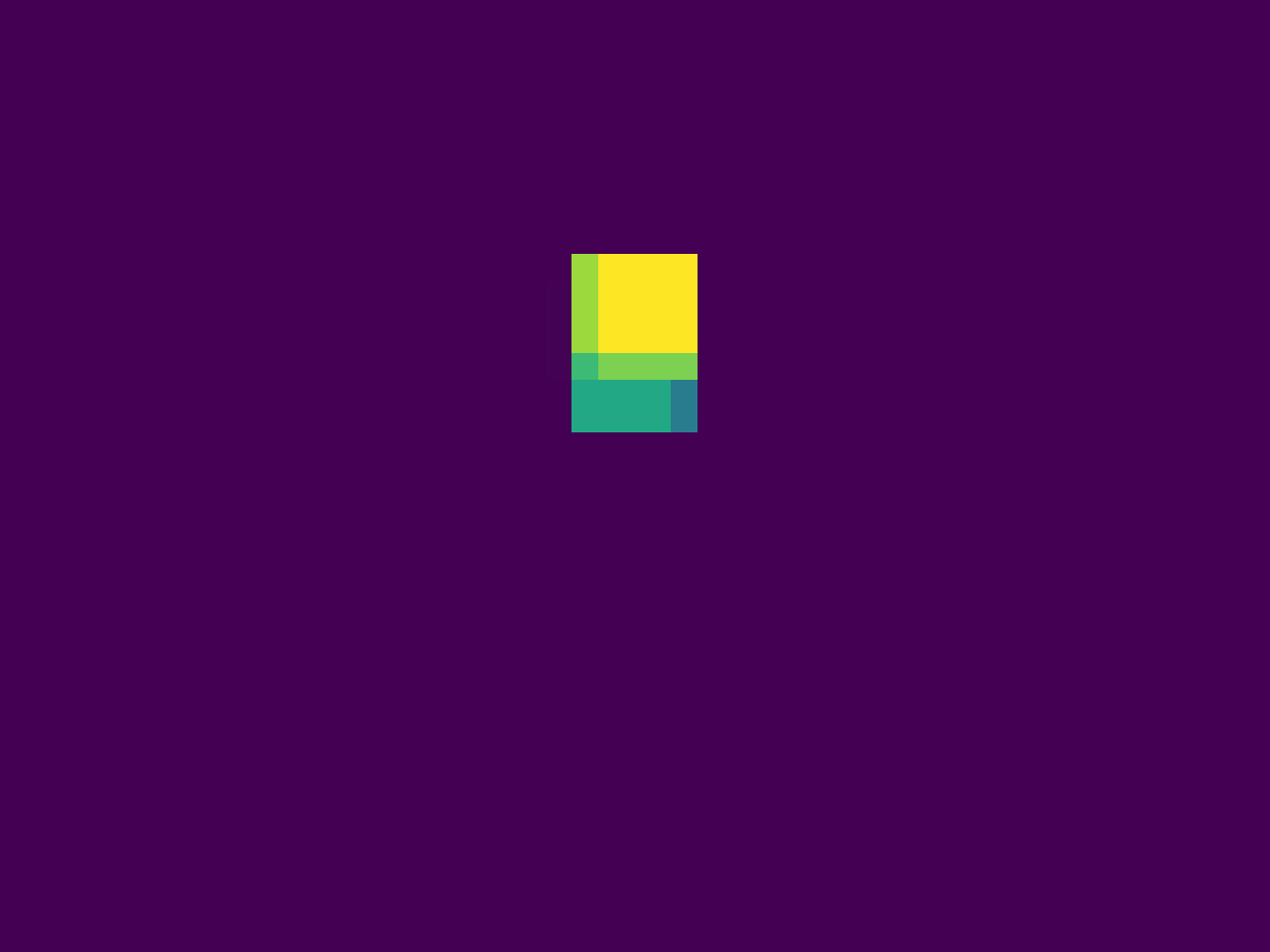}
        \caption{}
    \end{subfigure}
    \begin{minipage}[b]{0.11\textwidth}
        \begin{subfigure}[b]{\textwidth}
            \centering
            \includegraphics[width=\textwidth]{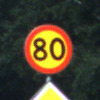}
            \caption{}
        \end{subfigure}
        \vspace{1em}
    \end{minipage}
    \caption{Visualization of the positions of the speed limit signs in test
             images of the dataset as well as the two attention distributions
             of \mil{} (left) and \ats{} (right) and the patches extracted from
             the high resolution image at the positions of the signs. Both
             methods identify effectively the speed limit in the high
             resolution image.}
    \label{fig:speed_limits}
\end{figure*}

Figure~\ref{fig:speed_limits} compares the attention distributions of \mil{}
and \ats{} on the Speed Limits dataset (\S~4.4 in the main paper). This dataset
is hard because it presents large variations in scale and orientation of the
regions of interest, namely the speed limit signs. However, we observe that
both methods locate effectively the signs even when there exist more than one
in the image. Note that for some of the images, such as
\subref{fig:speed_limits1} and \subref{fig:speed_limits2}, the sign is not
readable from the low resolution image.

\section{Network Architecture Details} \label{sec:networks}

In this section, we detail the network architectures used throughout our
experimental evaluation. The ultimate detail is always code, thus we encourage
the reader to refer to the github repository
\url{https://github.com/idiap/attention-sampling}.

\subsection{Megapixel MNIST} \label{sec:megamnist}

We summarize the details of the architectures used for the current experiment.
For ATS, we use a three layer convolutional network with 8 channels followed by
a ReLU activation as the attention network and a convolutional network inspired
from LeNet-1 \cite{lecun1995comparison} with 32 channels and a global
max-pooling as a last layer as the feature network. We also use an entropy
regularizer with weight $0.01$. The CNN baseline is a
ResNet-16 that starts with $32$ channels for convolutions and doubles them
after every two residual blocks.

We train all the networks with the Adam \cite{kingma2014adam} optimizer with a
fixed learning rate of $10^{-3}$ for 500 epochs.

\subsection{Histopathology images}

We summarize the details of the architecture used for the experiment on the H\&E
stained images. For ATS, we use a three layer convolutional network with 8
channels followed by ReLU non linearities as the attention network with an
entropy regularizer weight $0.01$. The feature network of is the same as the
one proposed by \cite{ilse18a}. Regarding, the CNN baseline, we use a
ResNet \cite{he2016deep} with 8 convolutional layers and 32 channels
instead.

We train all the networks for 30,000 gradient updates with the Adam
optimizer with learning rate $10^{-3}$. 

\subsection{Speed Limits}

We detail the network architectures used for the current experiment.
For \ats{}, we use an attention network that consists of four convolutions
followed by ReLU non-linearities starting with $8$ channels and
doubling them after each layer. Furthermore, we add a max pooling layer with
pool size $8$ at the end to reduce the sampling space and use an entropy
regularizer weight of $0.05$. The feature network of both our model
and \mil{} is a ResNet with $8$ layers and $32$ channels. The CNN baseline is a
ResNet-16 that starts with $32$ channels for convolutions and doubles them
after every two residual blocks.

Again, we we use the Adam \cite{kingma2014adam} optimizer with a
fixed learning rate of $10^{-3}$ for 300,000 iterations.

\end{document}